\def\paperID{*****} %
\def\confName{CVPR}
\def\confYear{2026}

\def\paperTitle{CleanStyle: Plug-and-Play Style Conditioning Purification \\ for Text-to-Image Stylization}

\def\authorBlock{
    Xiaoman Feng$^{*}$ \qquad
    Mingkun Lei$^{*,\dagger}$ \qquad
    Yang Wang \qquad
    Dingwen Fu \qquad
    Chi Zhang$^{\ddagger}$ \\
    AGI Lab, Westlake University \\
    {\footnotesize $^{*}$Equal contribution \quad $^{\dagger}$Project Leader \quad $^{\ddagger}$Corresponding author} \\
    \url{https://github.com/Westlake-AGI-Lab/CleanStyle}
}

\newif\ifreview 
\newif\ifarxiv \newcommand{\arxiv}{\arxivtrue}
\newif\ifcamera 
\newif\ifrebuttal 

\arxiv %

\pdfoutput=1
\documentclass[10pt,twocolumn,letterpaper]{article}
\ifreview \usepackage[review]{cvpr} \fi
\ifarxiv \usepackage[pagenumbers]{cvpr} \fi
\ifrebuttal \usepackage[rebuttal]{cvpr} \fi
\ifcamera \usepackage{cvpr} \fi

\usepackage{amsfonts,bm}

\usepackage{graphicx}	
\usepackage{amsmath}	
\usepackage{amssymb}	
\usepackage{booktabs}
\usepackage{times}
\usepackage{microtype}
\usepackage{epsfig}
\usepackage{caption}
\usepackage{float}
\usepackage{placeins}
\usepackage{color, colortbl}
\usepackage{stfloats}
\usepackage{enumitem}
\usepackage{tabularx}
\usepackage{xstring}
\usepackage{multirow}
\usepackage{xspace}
\usepackage{url}
\usepackage{subcaption}
\usepackage{xcolor}
\usepackage[hang,flushmargin]{footmisc}
\usepackage{algorithm}
\usepackage{algpseudocode}
\usepackage{wrapfig}
\usepackage{diagbox}
\usepackage{titletoc}

\ifcamera \usepackage[accsupp]{axessibility} \fi

\ifarxiv  \fi

\newcommand{\R}[1]{{%
    \textbf{%
        \ifstrequal{#1}{1}{\textcolor{red}{R#1}}{%
        \ifstrequal{#1}{2}{\textcolor{blue}{R#1}}{%
        \ifstrequal{#1}{3}{\textcolor{magenta}{R#1}}{%
        \ifstrequal{#1}{4}{\textcolor{teal}{R#1}}{%
                           \textcolor{cyan}{R#1}%
        }}}}%
    }%
}}

\def\eqref#1{equation~\ref{#1}}

\def\1{\bm{1}}

\DeclareMathAlphabet{\mathsfit}{\encodingdefault}{\sfdefault}{m}{sl}
\SetMathAlphabet{\mathsfit}{bold}{\encodingdefault}{\sfdefault}{bx}{n}

\usepackage{xr-hyper}

\makeatletter
\newcommand*{\addFileDependency}[1]{
  \typeout{(#1)}
  \@addtofilelist{#1}
  \IfFileExists{#1}{}{\typeout{No file #1.}}
}

\makeatother

\definecolor{cvprblue}{rgb}{0.21,0.49,0.74}
\usepackage[pagebackref,breaklinks,colorlinks,allcolors=cvprblue]{hyperref}
\usepackage[capitalize]{cleveref}
\crefname{section}{Sec.}{Secs.}
\crefname{table}{Table}{Tables}
\crefname{figure}{Fig.}{Figs.}

\ifarxiv \crefname{appendix}{App.}{Apps.}
\else \crefname{appendix}{Suppl.}{Suppls.} \fi

\begin{document}
\newcommand{\ours}{\texttt{CleanStyle}}

\newcommand{\tableCellHeight}{1}
\newcommand{\tabstyle}[1]{
  \setlength{\tabcolsep}{#1}
  \renewcommand{\arraystretch}{\tableCellHeight}
  \centering
  \small
}

\definecolor{tabhighlight}{HTML}{e5e5e5}
\definecolor{lightCyan}{rgb}{0.925,1,1}

\newtheorem{definition}{Definition}
\newtheorem{theorem}{Theorem}
\newtheorem{assumption}{Assumption}
\newtheorem{lemma}{Lemma}
\newtheorem{proposition}{Proposition}
\newtheorem{corollary}{Corollary}

\title{\paperTitle}
\author{\authorBlock}
\maketitle
\begin{abstract}
Style transfer in diffusion models enables controllable visual generation by injecting the style of a reference image. However, recent encoder-based methods, while efficient and tuning-free, often suffer from content leakage, where semantic elements from the style image undesirably appear in the output, impairing prompt fidelity and stylistic consistency.
In this work, we introduce \textbf{CleanStyle}, a plug-and-play framework that filters out content-related noise from the style embedding without retraining. Motivated by empirical analysis, we observe that such leakage predominantly stems from the tail components of the style embedding, which are isolated via Singular Value Decomposition (SVD). To address this, we propose \textbf{CleanStyleSVD (CS-SVD)}, which dynamically suppresses tail components using a time-aware exponential schedule, providing clean, style-preserving conditional embeddings throughout the denoising process.
Furthermore, we present \textbf{Style-Specific Classifier-Free Guidance (SS-CFG)}, which reuses the suppressed tail components to construct style-aware unconditional inputs.
Unlike conventional methods that use generic negative embeddings (\eg, zero vectors), SS-CFG introduces targeted negative signals that reflect style-specific but prompt-irrelevant visual elements.
This enables the model to effectively suppress these distracting patterns during generation, thereby improving prompt fidelity and enhancing the overall visual quality of stylized outputs.
Our approach is lightweight, interpretable, and can be seamlessly integrated into existing encoder-based diffusion models without retraining. Extensive experiments demonstrate that CleanStyle substantially reduces content leakage, improves stylization quality and improves prompt alignment across a wide range of style references and prompts.
\end{abstract}
    
\section{Introduction}
\label{sec:intro}
Recent advancements in text-to-image (T2I) generation have been driven by the rapid evolution of diffusion models~\cite{ho2020denoisingdiffusionprobabilisticmodels, song2022denoisingdiffusionimplicitmodels, rombach2022highresolutionimagesynthesislatent, podell2023sdxlimprovinglatentdiffusion, peebles2023scalablediffusionmodelstransformers}. These models, exemplified by Stable Diffusion (SD), have demonstrated remarkable capacity to synthesize high-quality images conditioned on textual prompts. Building on their strong generative priors, recent efforts have expanded the capabilities of T2I models to support fine-grained customization tasks, including image editing~\cite{hertz2022prompttopromptimageeditingcross, brooks2023instructpix2pixlearningfollowimage, flowedit}, personalized generation~\cite{ruiz2023dreamboothfinetuningtexttoimage, wang2024instantidzeroshotidentitypreservinggeneration, guo2024pulidpurelightningid, jiang2025infiniteyou}, and especially style transfer~\cite{wang2024instantstylefreelunchstylepreserving, gao2025styleshotsnapshotstyle, hertz2024stylealignedimagegeneration, qi2024deadiffefficientstylizationdiffusion, xu2025stylesspsamplingstartpointenhancement, lei2025stylestudio, xing2024csgocontentstylecompositiontexttoimage, omnistyle}.

\begin{figure}[t]
    \centering
    \includegraphics[width=1.0\linewidth]{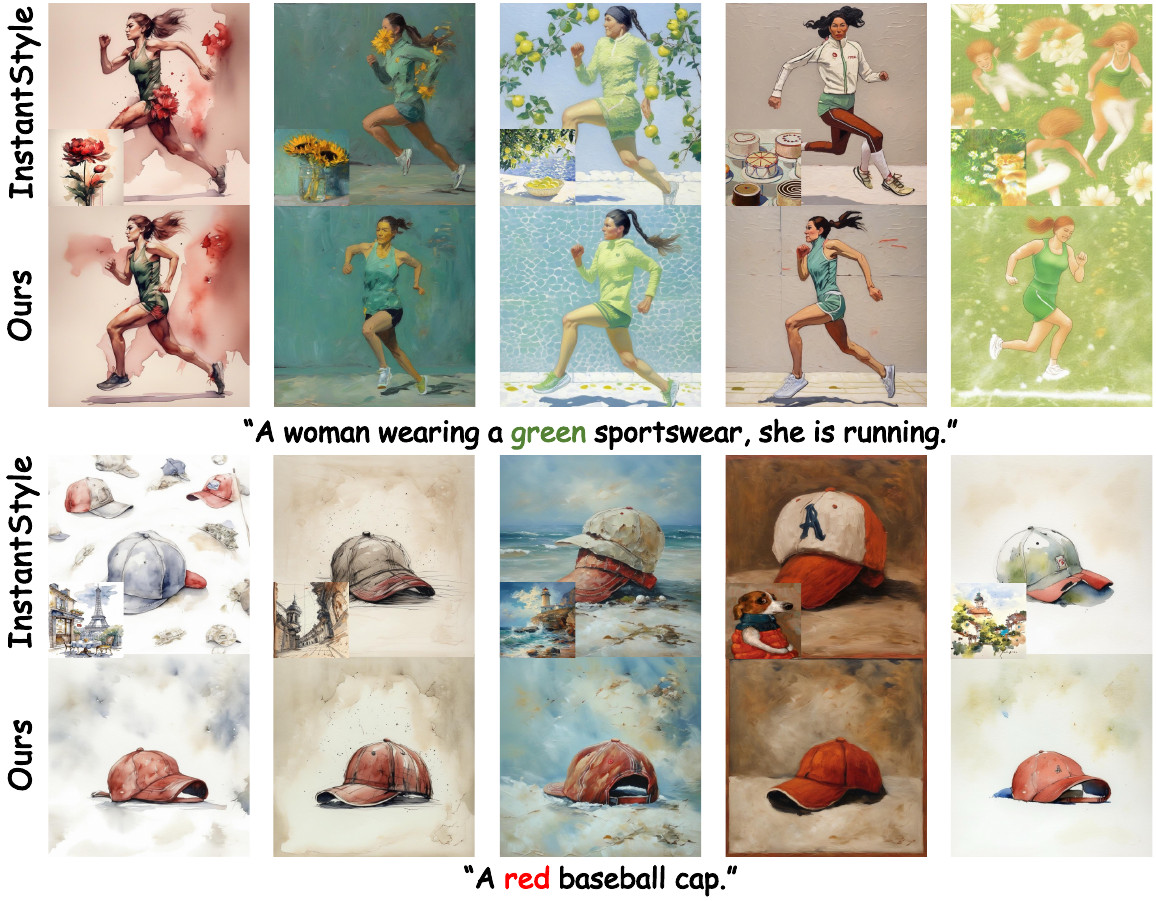}
    \caption{$\ours$ improves text-aligned style transfer by effectively mitigating content leakage. Compared to InstantStyle, our results better preserve prompt semantics while faithfully reflecting the reference style.}
    \label{fig:teaser}
    \vspace{-4mm}
\end{figure}

Among these, style transfer, injecting the visual style of a reference image into the generation process, has gained particular attention due to its applications in personalized content creation, design, and creative arts. The goal is to preserve the semantic alignment with the textual prompt while rendering the image with desired stylistic characteristics. To achieve this, encoder-based methods have emerged as a dominant paradigm~\cite{ye2023ipadaptertextcompatibleimage, wang2024instantstylefreelunchstylepreserving, hertz2024stylealignedimagegeneration}, owing to their feedforward design, fast inference, and compatibility with pretrained diffusion pipelines. These methods typically extract style embeddings from reference images using pretrained image encoders and inject them into cross-attention modules during sampling. By leveraging fixed encoders and compatible architectures, encoder-based frameworks achieve a favorable balance between flexibility, computational efficiency, and plug-and-play compatibility, making them particularly suitable for fast and scalable style transfer applications. 

A critical challenge in encoder-based style transfer is content leakage, where semantic details from the style reference are undesirably rendered in the final output. This phenomenon indicates that the extracted style representation is fundamentally an impure signal, conflating the desired, holistic stylistic attributes with undesired, content-specific information.
This suggests an analytical filtering mechanism is needed to decontaminate this signal.
We observe that this separation can be achieved by analyzing the embedding's singular spectrum via Singular Value Decomposition (SVD), a canonical method for signal component analysis. Our key insight is that a structural separation exists: the dominant, high-variance components encode the global style, while the low-variance tail components are the primary carriers of localized, content-specific artifacts.
Based on this premise, our core design is to analytically filter these tail components to mitigate content interference. To this end, we propose~\ours, a training-free and plug-and-play framework. Its central module, CleanStyleSVD~(CS-SVD), applies SVD to the style embeddings injected into the cross-attention layers. It then systematically suppresses the identified tail components using a time-aware exponential schedule. This dynamic approach applies stronger filtering during the early denoising steps, which are crucial for establishing a clean global layout, and progressively relaxes the suppression in later steps to preserve fine-grained stylistic details. This dynamic filtering process enables the model to retain stylistic expressiveness while substantially mitigating content contamination.

The generation process in modern diffusion models relies heavily on Classifier-Free Guidance (CFG), which functions by contrasting a positive (conditional) input against a negative (unconditional) one. However, the conventional CFG design is fundamentally ill-suited for this task, as it is ``style-agnostic". Standard methods employ generic negative inputs, such as zero vectors, which provide the model with no meaningful, style-relevant information to push against. This ``blind" guidance is inefficient: it tells the model what to become (the styled content), but offers no specific instruction on what to avoid.
This limitation motivated the design of our Style-Specific Classifier-Free Guidance (SS-CFG). We recognize that the components separated by CS-SVD, namely the tail components related to content leakage, are not just noise to be discarded. Instead, they could be repurposed to serve as a highly specific and targeted negative condition. SS-CFG, therefore, replaces the generic unconditional input with a style-aware negative embedding constructed directly from these tail components. This strategy establishes a precise contrastive objective: the model is guided to not only adhere to the ``clean" style (the filtered dominant components) but to actively diverge from the ``content-contaminated" signal (the tail components). This targeted negative guidance enables the model to more effectively suppress these confounding visual patterns, significantly enhancing prompt fidelity and the overall quality of the stylized output.

\begin{figure*}[t]
    \centering
    \includegraphics[width=1.0\linewidth]{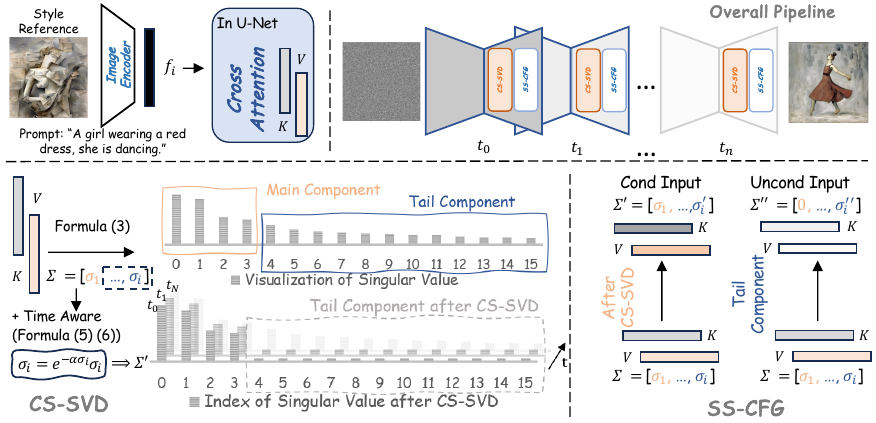}
    \caption{Overview of~$\ours$. We decompose cross-attention style embeddings via SVD into main and tail components, apply time-aware suppression to the tail component in CS-SVD, and form conditional embeddings. From the visualization of singular value (the Key $K$ is used as as an example), at the earlier time step $t_{0}$, suppression is stronger, while suppression is weaker at the later time step to preserve style details. SS-CFG uses the isolated tail component to build style-aware unconditional inputs. The figure shows the decomposition, the time-dependent filtering, and the conditional/unconditional pathways in sampling.}
    \label{fig:pipeline}
    \vspace{-4mm}
\end{figure*}

Our approach is general and modular: it can be integrated into a wide range of encoder-based methods (\eg, InstantStyle~\cite{wang2024instantstylefreelunchstylepreserving}, DEADiff~\cite{qi2024deadiffefficientstylizationdiffusion}, and StyleShot~\cite{gao2025styleshotsnapshotstyle}) with minimal changes and no retraining. When applied to these models, $\ours$ consistently improves generation quality by reducing content leakage and enhancing prompt adherence. The pipeline is presented in~\cref{fig:pipeline}.
Our main contributions are summarized as follows:
\begin{itemize}
  \item We conduct an empirical analysis of style embeddings and identify a key source of content leakage in encoder-based diffusion: tail components encode unintended semantic details from the reference image.
  \item We propose \textbf{CleanStyleSVD~(CS-SVD)}, a training-free filtering scheme that suppresses the tail components of style embeddings using a dynamic time-aware schedule.
  \item We introduce \textbf{Style-Specific Classifier-Free Guidance (SS-CFG)}, which reuses the suppressed components as unconditional inputs, replacing unspecific negatives and enabling stronger stylistic control.
  \item Our method is lightweight, interpretable, and broadly compatible with existing encoder-based diffusion models. Extensive experiments demonstrate significant reductions in content leakage and improvements in stylization quality. Codes will be publicly released to facilitate future research.
\end{itemize}

\section{Related work}
\subsection{Style transfer with Diffusion Models}
Style transfer aims to render a target image in the visual appearance of a reference image. With the rapid advancement of text-to-image diffusion models~\cite{rombach2022highresolutionimagesynthesislatent, podell2023sdxlimprovinglatentdiffusion, esser2024scalingrectifiedflowtransformers}, recent methods~\cite{zhou2025attentiondistillationunifiedapproach, nguyen2025csd, lei2025stylestudio, xu2025stylesspsamplingstartpointenhancement} have focused on improving the fidelity, flexibility, and controllability of stylized generation.

Encoder-based methods leverage pre-trained image encoders (\eg, CLIP~\cite{radford2021clip}) to extract style representations and inject them into the diffusion model. For instance, IP-Adapter~\cite{ye2023ipadaptertextcompatibleimage} aligns image features with textual prompts to enable visual conditioning. InstantStyle~\cite{wang2024instantstylefreelunchstylepreserving} builds upon IP-Adapter by selectively injecting style features into specific U-Net~\cite{ronneberger2015u} layers. StyleShot~\cite{gao2025styleshotsnapshotstyle} proposes a multi-scale encoder to capture fine-grained and global style elements. CSGO~\cite{xing2024csgocontentstylecompositiontexttoimage} constructs a dedicated dataset to supervise the disentanglement of style and content. DEADiff~\cite{qi2024deadiffefficientstylizationdiffusion} introduces a joint text-image cross-attention layer to enhance prompt adherence in stylized generation. These methods collectively aim to address a core challenge in encoder-based style transfer: content leakage from the style reference, which can compromise prompt fidelity and visual coherence.

Other works rely on model fine-tuning, inversion, or parameter-efficient adaptation. InST~\cite{zhang2023inversionbasedstyletransferdiffusion} optimizes latent codes for style reconstruction. StyleDrop~\cite{sohn2023styledrop} introduces iterative fine-tuning with feedback to refine stylization. DreamBooth~\cite{ruiz2023dreamboothfinetuningtexttoimage} enforces subject identity through prior preservation. DreamStyler~\cite{ahn2024dreamstyler} introduces prompt-level augmentation to decouple visual style from implicit content. LoRA-based methods, such as B-LoRA~\cite{frenkel2024implicitstylecontentseparationusing}, K-LoRA~\cite{ouyang2025k}, ZipLoRA~\cite{shah2023ziplorasubjectstyleeffectively}, and UnzipLoRA~\cite{liu2024unziplora}, aim to disentangle or recombine content and style by fine-tuning low-rank adapters. StyleAlign~\cite{hertz2024stylealignedimagegeneration} swaps query-key positions in attention to better align style semantics, StyleKeeper~\cite{stylekeeper} extends prior style-alignment designs by introducing a CFG-based neg-style guidance construction.

Our work builds upon encoder-based pipelines and tackles content leakage from a new perspective: we introduce a plug-and-play method that analytically filters style embeddings via SVD, requiring no training or fine-tuning. This offers a simple yet effective alternative to disentanglement-based training strategies.

\subsection{Singular Value Decomposition (SVD) in Diffusion Models}
SVD has recently been applied in diffusion models for diverse purposes such as model compression, feature filtering, and controllable generation. SVDiff~\cite{han2023svdiffcompactparameterspace} performs fine-tuning in the singular value space to reduce parameter count. 1Prompt1Story~\cite{liu2025onepromptonestoryfreelunchconsistenttexttoimage} decomposes text embeddings to amplify or suppress narrative elements. Get What You Want~\cite{li2024get} applies SVD to isolate undesirable text concepts from conditioning embeddings.

In contrast, our work is the first to apply SVD to image-based style embeddings for stylization. Rather than manipulating text embeddings, we analyze the singular spectrum of style embedding and empirically observe that tail components tend to correlate with content-specific artifacts. This motivates a new application of SVD in filtering residual content signals from style embeddings in diffusion-based generation.

\section{Method}
\subsection{Preliminaries}
\noindent\textbf{Encoder-based style transfer.}
Recent diffusion-based style transfer methods~\cite{wang2024instantstylefreelunchstylepreserving, qi2024deadiffefficientstylizationdiffusion, xing2024csgocontentstylecompositiontexttoimage, gao2025styleshotsnapshotstyle} extract style embeddings using image encoders and inject them into the U-Net via the Key and Value branches of cross-attention layers.

Concretely, a reference style image $I_s$ is first encoded into a feature vector $e_s$, which is then projected by a trainable MLP and transformed into Key ($K$) and Value ($V$) representations via learned matrices $W_K$ and $W_V$. These representations are injected at each denoising step, enabling the network to incorporate visual style cues alongside the text prompt.
Building on this injection mechanism, we introduce a plug-and-play filtering module to suppress content leakage, ensuring better prompt consistency and overall generation quality, as detailed in the following sections.

\noindent\textbf{Singular Value Decomposition (SVD).}
Given a real matrix $X \in \mathbb{R}^{m \times n}$, its singular value decomposition is:
\begin{equation}
X = U \Sigma V^{\top},
\end{equation}
where $U \in \mathbb{R}^{m \times r}$ and $V \in \mathbb{R}^{n \times r}$ are orthogonal matrices, and $\Sigma = \mathrm{diag}(\sigma_1, \dots, \sigma_r)$ is a diagonal matrix with non-increasing singular values $\sigma_1 \geq \dots \geq \sigma_r > 0$, and $r = \mathrm{rank}(X)$.
The top singular values often encode the most significant variance directions in the data, whereas the remaining components are associated with less informative or noisy variations.

\noindent\textbf{Classifier-Free Guidance (CFG).}
CFG~\cite{ho2022classifierfreediffusionguidance, karras2024guidingdiffusionmodelbad} is a widely adopted strategy in diffusion models for amplifying the influence of conditional signals during generation. It interpolates between the conditional and unconditional noise predictions as follows:
\begin{equation}
\boldsymbol{\epsilon}_{\text{CFG}} = \boldsymbol{\epsilon}_{\text{uncond}} + \omega \cdot (\boldsymbol{\epsilon}_{\text{cond}} - \boldsymbol{\epsilon}_{\text{uncond}})
\label{eq:cfg}
\end{equation}
where $\boldsymbol{\epsilon}_{\text{cond}}$ and $\boldsymbol{\epsilon}_{\text{uncond}}$ denote the predicted noises with and without conditioning, and $\omega$ is the guidance scale. 

In encoder-based style transfer pipelines, the unconditional branch is typically formed by feeding a null embedding (\eg, zero vectors) as the style input. However, this yields a generic and style-agnostic signal, which limits the model’s ability to distinguish style-specific features from content leakage.
We later leverage the empirical observation that tail components tend to encode content-related information to design a semantically meaningful unconditional branch.

\begin{figure}[t]
    \centering
    \includegraphics[width=1.0\linewidth]{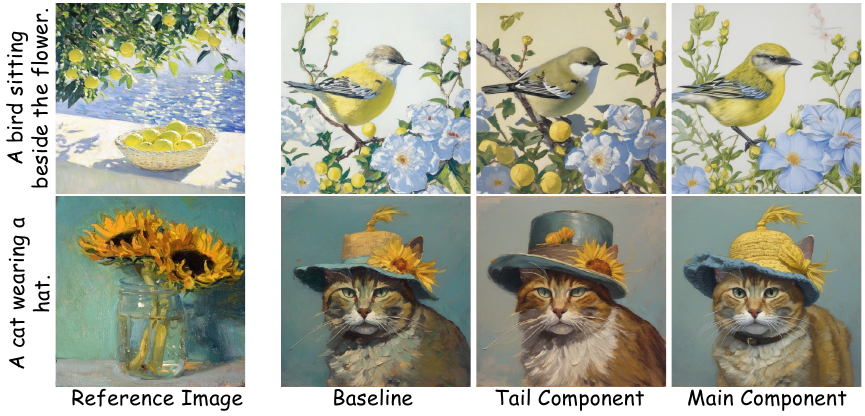}
    \caption{Motivational illustration. The baseline exhibits clear content leakage. Using only the \textit{tail component} (as defined in ~\cref{fig:pipeline}) further amplifies these artifacts, indicating that the tail region mainly encodes content-related signals rather than stylistic information. Conversely, relying solely on the \textit{main component} weakens the overall style expression. These observations motivate our design: CS-SVD suppresses tail-induced content leakage, while the time-aware strategy modulates this suppression to avoid over-attenuating stylistic details, achieving a balanced and faithful stylization.}
    \label{fig:motivation}
    \vspace{-6mm}
\end{figure}

\subsection{CleanStyleSVD}
\label{sec:svd-filter}
In text-to-image style transfer, we observe that style embeddings often contain residual content signals, which may compromise prompt alignment and visual coherence. Empirical analysis (see~\cref{fig:motivation}) reveals that such content-relevant information is primarily captured by the tail components of the embedding’s singular spectrum, and is largely irrelevant to style. Based on this observation, we propose to apply SVD as a filtering mechanism to suppress tail components and reduce content leakage during generation.

Formally, let $f \in \mathbb{R}^{d \times N}$ denote the flattened feature map from a style image, where $d$ is the feature dimension and $N$ is the number of spatial tokens. In encoder-based pipelines, the feature is first projected via a learned weight matrix $W_K$ into the cross-attention Key:

\begin{equation}
K = W_K f = U \Sigma V^\top,
\end{equation}
where $U \in \mathbb{R}^{d \times d}$, $V \in \mathbb{R}^{N \times N}$ are orthogonal matrices and $\Sigma \in \mathbb{R}^{d \times N}$ is a rectangular diagonal matrix containing singular values $\{\sigma_1, \sigma_2, \dots, \sigma_r\}$ with $\sigma_1 \geq \dots \geq \sigma_r$. A similar decomposition is applied to the Value matrix. Instead of filtering the encoder feature $f$, we apply SVD to the projected Key and Value matrices. As shown in~\cref{fig:abl_timeaware}, filtering $f$ alone yields insufficient suppression, while our approach achieves stronger content removal. This validates that operating at the attention-level representation is more effective.

To suppress tail signals associated with content artifacts, we attenuate the singular values beyond the top-$k$ components using exponential decay:
\begin{equation}
\sigma_i' =
\begin{cases}
\sigma_i, & \text{if } i \leq k \\
e^{-\alpha \sigma_i} \cdot \sigma_i, & \text{otherwise}
\end{cases}
\end{equation}
where $\alpha$ is a suppression factor controlling the decay strength. 
Under this formulation, larger singular values within the tail region undergo stronger attenuation, allowing the method to selectively damp prominent content-related signals while preserving dominant stylistic information.

To adapt to the denoising dynamics, we further introduce a time-dependent suppression schedule. Following prior work~\cite{agarwal2023imageworthmultiplewords, hertz2022prompttopromptimageeditingcross, brooks2023instructpix2pixlearningfollowimage} showing that layout and structure are determined in early denoising steps, we apply stronger suppression early on and gradually relax it:
\begin{align}
s(t) &= \frac{1}{1 + e^{-\gamma \left( \frac{t}{T} - c \right)}}, \\
\alpha_t &= \alpha_0 \cdot (1 - s(t)),
\end{align}
where $T$ is the total number of denoising steps, and $\gamma, c$ control the steepness and midpoint of the schedule. This formulation guarantees that $\alpha_t$ decreases as the denoising step $t$ increases, providing strong suppression when global structure is formed and progressively reducing its effect to preserve fine-grained style details. As shown in~\cref{fig:abl_timeaware}, this time-aware strategy is crucial for preserving fine-grained stylistic details such as brush strokes and color tones. Without it, the style may become overly faded or homogenized.

This SVD-based filtering module is a plug-and-play, training-free mechanism that improves prompt fidelity and suppresses content leakage. The method is lightweight and introduces negligible inference overhead, making it suitable for integration into existing diffusion models.

\subsection{Style-Specific CFG}
\label{sec:ss-cfg}
Classifier-Free Guidance (CFG)~\cite{ho2022classifierfreediffusionguidance} is a standard mechanism for strengthening prompt alignment in diffusion models. However, when applied to encoder-based style transfer, existing approaches typically set the unconditional branch to a zero or generic embedding, which does not capture the instance-specific residual signals present in the style embedding. As a result, the guidance term may fail to suppress style-specific but prompt-irrelevant content, leading to weakened prompt fidelity or unintended visual artifacts.

To address this limitation, we introduce \textbf{Style-Specific CFG (SS-CFG)}, which constructs a style-aware unconditional embedding directly from the \emph{tail component} of the style representation. After applying CS-SVD, the style embedding is decomposed into a filtered component that preserves dominant stylistic cues and a tail component that retains prompt-irrelevant or content-related signals. SS-CFG assigns these components to the two guidance branches as follows:
\begin{itemize}
    \item \textbf{Conditional branch ($\epsilon_{\text{cond}}$):} uses the CS-SVD filtered Key/Value embeddings, where tail singular values are attenuated by the time-aware schedule.
    \item \textbf{Unconditional branch ($\epsilon_{\text{uncond}}$):} uses the isolated tail component to provide a targeted negative signal that captures undesired content tendencies specific to the style image.
\end{itemize}

This contrastive construction preserves compatibility with the standard CFG mechanism while introducing a style-aware unconditional pathway. By replacing generic unconditional embeddings with an instance-specific negative signal, SS-CFG more effectively suppresses unwanted content features and yields stronger, more stable prompt alignment during generation.
Our method is training-free, fully compatible with existing encoder-based style transfer pipelines, and significantly improves prompt fidelity by explicitly modeling negative cues that are otherwise ignored in conventional CFG formulations.

\subsection{Integration with other SOTAs}
\label{integrated}

To validate the generality of our method, we integrate~$\ours$ into two representative encoder-based stylization frameworks: StyleShot~\cite{gao2025styleshotsnapshotstyle} and DEADiff~\cite{qi2024deadiffefficientstylizationdiffusion}.

Although StyleShot~\cite{gao2025styleshotsnapshotstyle} introduces a multi-scale encoder to enrich style representation, its style injection mechanism remains Adapter-based~\cite{ye2023ipadaptertextcompatibleimage}, where style features are incorporated through cross-attention. Consequently, our method can be applied to StyleShot in the same manner as in InstantStyle~\cite{wang2024instantstylefreelunchstylepreserving}, by operating directly on the style-related Key and Value matrices in each attention layer.

DEADiff~\cite{qi2024deadiffefficientstylizationdiffusion} adopts a joint text–image attention design in which text and style embeddings are concatenated at the attention level:
\begin{equation}
A = \text{Softmax}\!\left( \frac{Q [K_t \, ; \, K_i]^T}{\sqrt{d}} \right) [V_t \, ; \, V_i].
\end{equation}
Within this structure, we apply~$\ours$ exclusively to the style components $K_i$ and $V_i$, leaving the text pathway unchanged. 

Across both frameworks, $\ours$ functions as a fully plug-and-play module that requires no retraining and introduces no architectural modifications, demonstrating its broad compatibility with existing encoder-based stylization pipelines.

\section{Evaluation and experiments}
\noindent \textbf{Implementation and evaluation setup.} We apply our method primarily to InstantStyle~\cite{wang2024instantstylefreelunchstylepreserving}, with additional adaptations to DEADiff~\cite{qi2024deadiffefficientstylizationdiffusion} and StyleShot~\cite{gao2025styleshotsnapshotstyle} to demonstrate generalizability. All evaluations are conducted under official inference settings without additional training.
Our primary experiments on InstantStyle~\cite{wang2024instantstylefreelunchstylepreserving} adopt the following hyperparameters: SVD truncation rank $k=1$, suppression factor $\alpha=0.01$, dynamic schedule parameters $\gamma=40$, $c=0.25$, and SS-CFG weight $w=5$. All experiments are conducted on a single RTX 4090 GPU.

\subsection{Comparisons with State-of-the-Arts}
We compare our method with encoder-based approaches, including InstantStyle~\cite{wang2024instantstylefreelunchstylepreserving}, StyleStudio~\cite{lei2025stylestudio}, DEADiff~\cite{qi2024deadiffefficientstylizationdiffusion}, StyleShot~\cite{gao2025styleshotsnapshotstyle}, CSGO~\cite{xing2024csgocontentstylecompositiontexttoimage}, and IP-Adapter~\cite{ye2023ipadaptertextcompatibleimage}. We use the same random seed for generation and fix the number of inference steps to 30 in all experiments.

\begin{figure*}[t]
    \centering
    \includegraphics[width=1.0\linewidth]{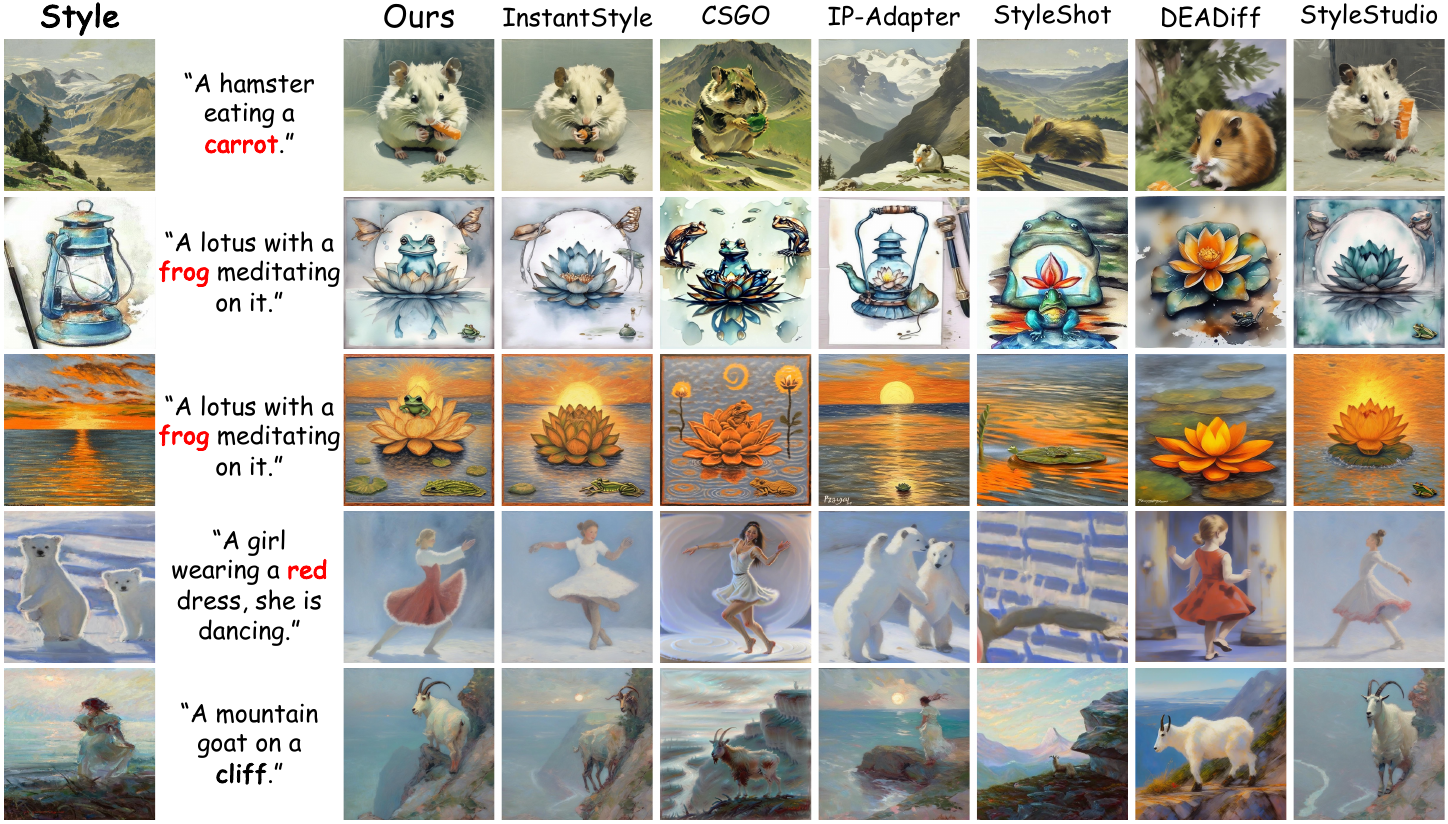}
    \caption{Qualitative comparison with the state-of-the-art encoder-based style transfer methods. Our approach effectively suppresses content leakage (row 1), achieves stronger prompt alignment (rows 2--4), 
    and maintains higher visual fidelity with fewer structural or stylistic distortions (row 5).}
    \label{fig:quality_main}
    \vspace{-4mm}
\end{figure*}

\noindent \textbf{Qualitative comparisons.} 
As shown in~\cref{fig:quality_main}, our method delivers consistent improvements across three major dimensions: mitigating content leakage, strengthening prompt alignment, and preserving visual quality. In the first row, existing methods such as CSGO~\cite{xing2024csgocontentstylecompositiontexttoimage} and DEADiff~\cite{qi2024deadiffefficientstylizationdiffusion} exhibit pronounced content leakage. In contrast, our result cleanly isolates style from content while faithfully rendering the target object.
Rows two through four highlight prompt alignment. Several baselines struggle to follow the textual instruction: InstantStyle~\cite{wang2024instantstylefreelunchstylepreserving} and StyleStudio~\cite{lei2025stylestudio} often produce incorrect object compositions or fail to realize the specified actions, while StyleShot~\cite{gao2025styleshotsnapshotstyle} may distort key semantic details. Our method accurately captures the intended scene (\eg, a frog meditating on a lotus) without compromising stylistic fidelity.
In the fifth row, we observe noticeable degradation in visual quality(\eg, structural inconsistencies) for multiple baselines, , or style corruption.
Additional stylization results and more diverse style exemplars are provided in the Appendix.

\begin{table}[t]
\centering
\setlength{\tabcolsep}{3pt}      %
\renewcommand{\arraystretch}{0.95}
\begin{tabular}{@{}lccc@{}}
\toprule
Method & Text Align$\uparrow$ & Style Sim$\uparrow$ & IQ$\uparrow$ \\
\midrule
\rowcolor{lightCyan}\textbf{InstantStyle$^\dag$} & \textbf{0.572} & \textbf{0.565} & \textbf{0.541} \\
InstantStyle & 0.133 & 0.128 & 0.128 \\
CSGO & 0.035 & 0.030 & 0.040 \\
IP-Adapter & 0.016 & 0.014 & 0.016 \\
StyleShot & 0.053 & 0.058 & 0.056 \\
DEADiff & 0.037 & 0.034 & 0.033 \\
StyleStudio & 0.153 & 0.171 & 0.185 \\
\bottomrule
\end{tabular}
\caption{User Study. IQ denotes Image Quality and $\dag$: Baseline integrated with our method.}
\label{tab:User Study}
\vspace{-4mm}
\end{table}

\noindent\textbf{User study.}
We collected 43 complete responses, totaling 2,580 judgments (43 participants $\times$ 60 questions)~\cref{tab:User Study}.
The user study consisting of 60 questions derived from 20 diverse pairs, providing broader coverage than the evaluation setups used in prior encoder-based stylization works, including DEADiff~\cite{qi2024deadiffefficientstylizationdiffusion}. Each pair is evaluated under three criteria: text alignment, style similarity, and overall image quality. Participants were shown outputs from all methods in randomized order and asked to select the best result for each criterion.

Across this expanded evaluation set, our method receives the highest overall preference, demonstrating stronger prompt adherence, better style retention, and superior perceptual quality. The full questionnaire interface is provided in the Appendix.

\begin{table}[t]
\small
\centering
\setlength{\tabcolsep}{3pt}      %
\renewcommand{\arraystretch}{0.95}
\begin{tabular}{lcccc}
\toprule
Method&CLIP TA$\uparrow$&CLIP SS$\uparrow$ &DINO SS$\uparrow$ &CT(s)\\
\midrule
\cellcolor{lightCyan}\textbf{InstantStyle$^\dag$} & \cellcolor{lightCyan}\textbf{0.231} & \cellcolor{lightCyan}0.711 & \cellcolor{lightCyan}0.386 & \cellcolor{lightCyan}4.81\\
InstantStyle & 0.228 & 0.718 & 0.396 & 4.13\\
\midrule
\cellcolor{lightCyan}\textbf{DEADiff$^\dag$} & \cellcolor{lightCyan}\textbf{0.245} & \cellcolor{lightCyan}\textbf{0.671} & \cellcolor{lightCyan}0.281 & \cellcolor{lightCyan}6.81\\  
DEADiff & 0.239 & 0.670 & 0.307 & 3.92\\
\midrule
\cellcolor{lightCyan}\textbf{StyleShot$^\dag$} & \cellcolor{lightCyan}\textbf{0.227} & \cellcolor{lightCyan}0.701 & \cellcolor{lightCyan}0.386 & \cellcolor{lightCyan}5.47\\
StyleShot & 0.221 & 0.717 & 0.431 & 4.23\\
\midrule
StyleStudio & 0.243 & 0.686 & 0.340 & 16.86\\
CSGO & 0.217 & 0.684 & 0.392 & 6.65\\
IP-Adapter & 0.135 & 0.885 & 0.675 & 4.69\\
\bottomrule
\end{tabular}
\caption{Quantitative Comparisons on~$\ours$. CT(s) means computation time in seconds. As shown, integrating our modules introduces only marginal overhead, resulting in computation times comparable to the original baselines. $\dag$: Baseline integrated with our method.}
\label{tab:quantitative_CleanStyle}
\vspace{-4mm}
\end{table}

\noindent \textbf{Quantitative comparisons.}
To evaluate the effectiveness and robustness of our method, we compare it against several encoder-based style transfer approaches. In addition, we integrate our method into InstantStyle(SDXL)~\cite{wang2024instantstylefreelunchstylepreserving}, DEADiff(SD1.5)~\cite{qi2024deadiffefficientstylizationdiffusion}, and StyleShot(SD1.5)~\cite{gao2025styleshotsnapshotstyle} to demonstrate its generalizability across architectures.
We evaluate on two datasets. StyleBench~\cite{gao2025styleshotsnapshotstyle} is a benchmark comprising 490 style images and 20 prompts. $\ours$ is a curated dataset constructed from 100 publicly available style images and 52 prompts adapted from StyleAdapter~\cite{wang2024styleadapterunifiedstylizedimage}, aiming to evaluate model generalization across a diverse range of artistic styles. The prompts span a wide range of semantic complexity, from simple object descriptions to multi-clause scene instructions, providing a comprehensive testbed for assessing both prompt alignment and stylistic consistency. The full set of style references used in $\ours$ is included in the Appendix.
For evaluation, we adopt three widely used metrics: CLIP Text Alignment(TA), and CLIP Style Similarity(SS)~\cite{radford2021clip} and DINO Style Similarity(SS)~\cite{caron2021dino}.

As shown in~\cref{tab:quantitative_CleanStyle}, our method consistently achieves superior CLIP-TA scores, indicating stronger prompt alignment. Compared to each method’s original baseline, our CLIP-SS and DINO-SS scores are slightly lower. This is expected, as both metrics partially rely on semantic features extracted by pre-trained encoders like CLIP~\cite{radford2021clip} and DINO~\cite{caron2021dino}. This phenomenon highlights an inherent trade-off in style transfer tasks. Our method achieves a more favorable balance between semantic consistency and style preservation. The higher computational overhead observed on DEADiff primarily arises from its larger style-related Key/Value tensors, which lead to a more expensive SVD decomposition compared to other architectures. Results on the StyleBench are provided in the Appendix.

\subsection{Further analysis.}

\begin{figure}[t]
    \centering
    \includegraphics[width=1.0\linewidth]{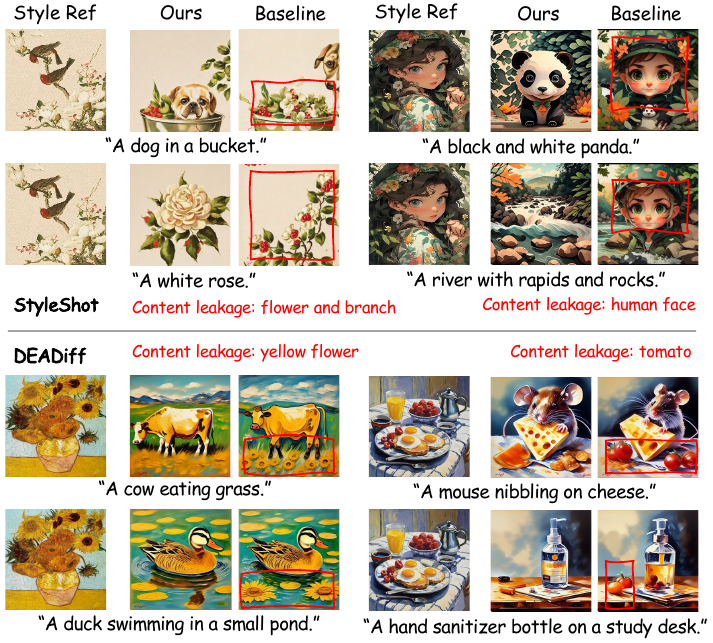}
    \caption{Integrated with StyleShot and DEADiff. On both the comparisons, ours mitigate the content leakage issue and keep stylistic features.}
    \label{fig:combine}
    \vspace{-4mm}
\end{figure}

\noindent \textbf{Integrated with other SOTAs.} To illustrate the effectiveness and generality of our approach, we integrate it with StyleShot~\cite{gao2025styleshotsnapshotstyle} and DEADiff~\cite{qi2024deadiffefficientstylizationdiffusion}. The results are in~\cref{fig:combine}. It is shown that on both the baselines, we mitigate content leakage such as the ``flower and branches", ``human face" and ``tomato". Also, our method keeps the stylistic characteristics such as color, texture and stroke. 
These results highlight the versatility of our method as a training-free, modular enhancement to existing style transfer frameworks.

\begin{figure}[t]
    \centering
    \includegraphics[width=1.0\linewidth]{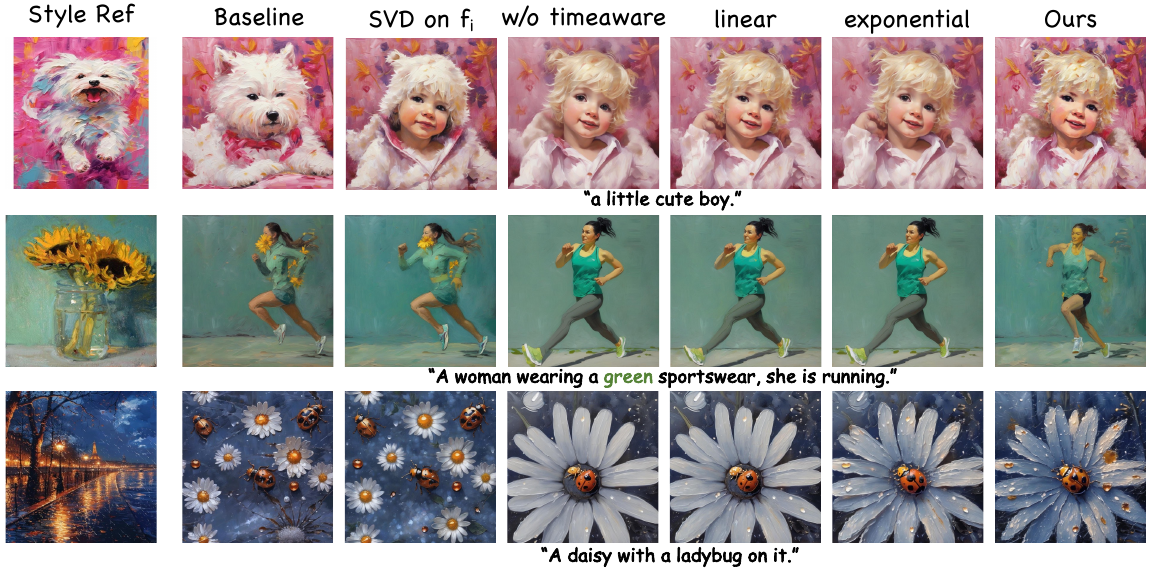}
    \caption{Comparison of different design choices. ``SVD on $f_i$" denotes filtering applied directly to the image encoder output. We evaluate four strategies: fixed (w/o time-aware), linear ($\alpha_t = \alpha_0 (1 - \frac{t}{T})$), exponential ($\alpha_t = \alpha_0 e^{-\lambda \frac{t}{T}}$), and ours. Our sigmoid-based schedule leads to stronger stylization and achieves the richest stylistic detail.}
    \label{fig:abl_timeaware}
    \vspace{-4mm}
\end{figure}

\begin{figure}[t]
    \centering
    \includegraphics[width=1.0\linewidth]{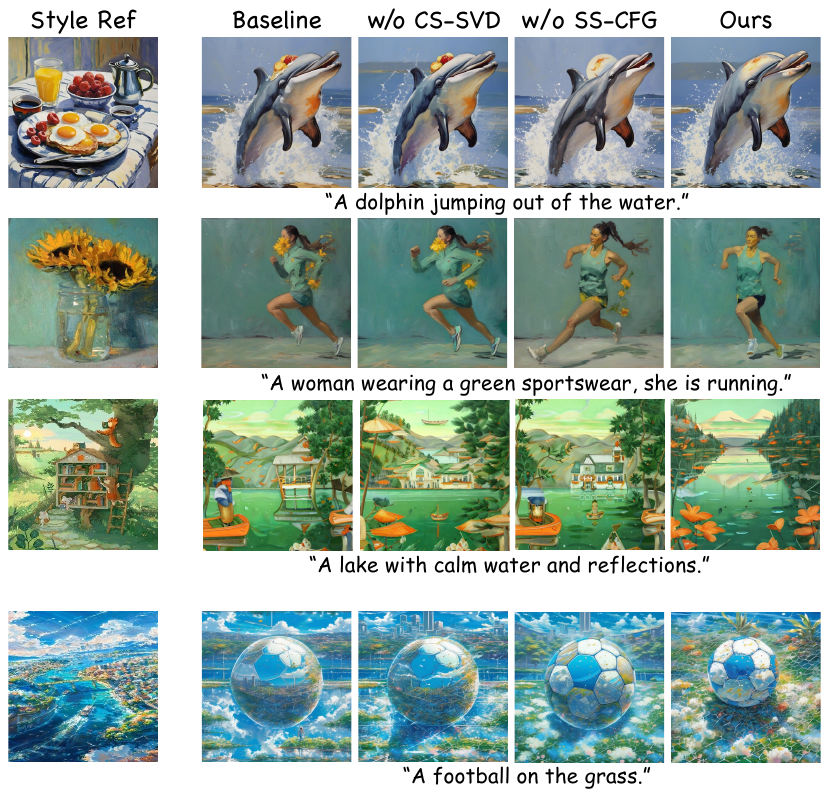}
    \caption{Qualitative ablation study of~$\ours$. Using SS-CFG alone (w/o CS-SVD, third column) produces outputs visually close to the baseline, indicating that the unconditional pathway cannot function effectively without a properly filtered conditional branch; this observation is also supported by the quantitative results in~\cref{tab:Ablation}. Using CS-SVD alone (w/o SS-CFG, fourth column) suppresses major artifacts but still leaves subtle style-related structures (\eg, flowers) that are difficult to remove. Only the full model eliminates content-leakage artifacts while maintaining strong prompt alignment, demonstrating that CS-SVD and SS-CFG are both necessary and complementary.}
    \label{fig:abl_components}
    \vspace{-4mm}
\end{figure}

\begin{table}[t]
\centering
\begin{tabular}{lcccc}
\toprule
Method&CLIP TA$\uparrow$ & CLIP SS$\uparrow$ & DINO SS$\uparrow$\\
\midrule
Baseline & 0.228 & \textbf{0.718} & \textbf{0.396}\\
w/o CS-SVD & 0.229 & \underline{0.717} & \underline{0.393}\\
w/o SS-CFG & \textbf{0.244} & 0.671 & 0.390\\
\rowcolor{lightCyan}Ours & \underline{0.231} & 0.711 & 0.386\\
\bottomrule
\end{tabular}
\caption{Quantitative ablation study of~$\ours$. Using SS-CFG alone offers limited improvement, as CFG requires a properly filtered conditional branch to take effect. Using CS-SVD alone improves prompt alignment but noticeably degrades style similarity, reflecting the inherent trade-off between content suppression and style preservation. Combining both modules achieves the most balanced performance across all metrics.}
\label{tab:Ablation}
\vspace{-4mm}
\end{table}

\begin{figure}[t]
    \centering
    \includegraphics[width=1.0\linewidth]{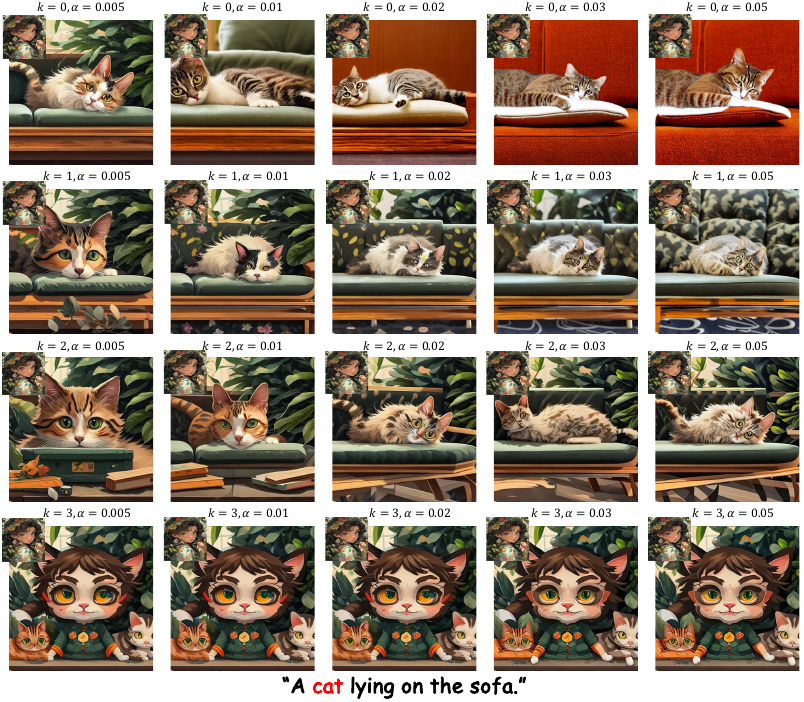}
    \caption{Comparison of different top-k value and $\alpha_0$. Larger $\alpha_0$ values apply stronger attenuation to tail components, leading to reduced stylistic influence. Conversely, increasing top-$k$ preserves more dominant singular components, which can reintroduce content leakage from the style image. A moderate setting achieves the best balance between style strength and content cleanliness.}
    \label{fig:abl_k_alpha}
    \vspace{-4mm}
\end{figure}

\begin{figure}[t]
    \centering
    \includegraphics[width=1.0\linewidth]{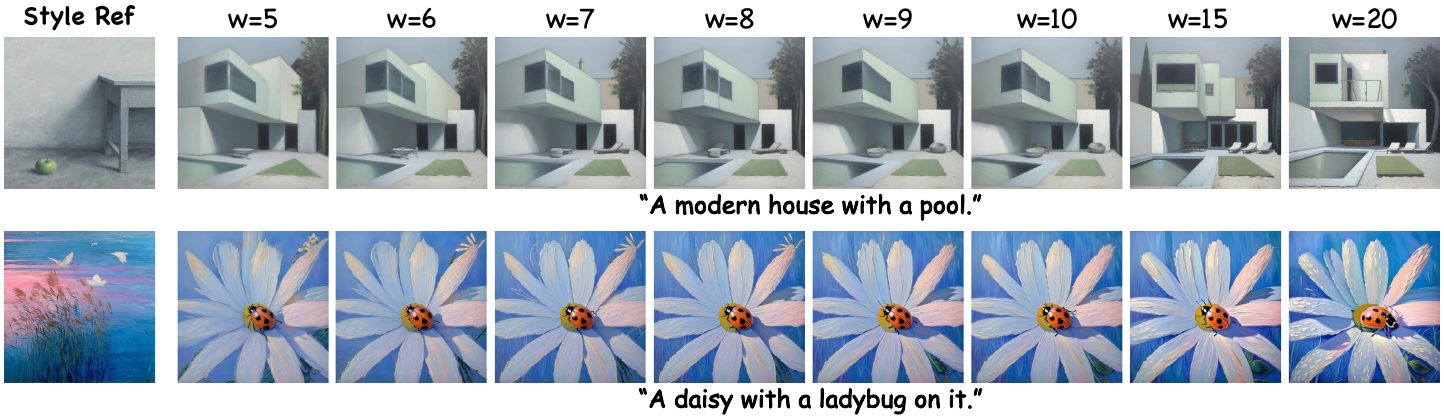}
    \caption{Effect of varying guidance weight $w$ in SS-CFG. SS-CFG maintains stable color distribution, coherent structure, and consistent style expression even at high $w$.}
    \label{fig:abl_sscfg_w}
    \vspace{-4mm}
\end{figure}

\noindent \textbf{Ablation study.}
Both components of~$\ours$ contribute critically to the final performance. As illustrated qualitatively in~\cref{fig:abl_components}, using SS-CFG alone produces outputs visually close to the baseline, whereas using CS-SVD alone leaves subtle style-related artifacts that are difficult to suppress. The quantitative results in~\cref{tab:Ablation} further corroborate these findings: CS-SVD is essential for removing content-related interference, while SS-CFG plays a key role in maintaining prompt alignment. Together, they provide complementary benefits and yield the best balance between semantic fidelity and style preservation.

\noindent \textbf{Design choice analysis.} We further analyze two critical design choices in our method. First, we compare applying SVD filtering on the intermediate Key/Value features against directly processing the image encoder output $f_i$. As shown in~\cref{fig:abl_timeaware}, filtering $f_i$ yields suboptimal results, often failing to suppress style-induced content leakage such as the dog's fur and the sunflower petals. Second, we assess the impact of different suppression schedules for tail components. Without timestep awareness, the results suffer from missing stylistic textures. Fixed strategies such as linear or exponential suppression only offer limited improvement. 

\noindent \textbf{Hyperparameter analysis.} We further investigate the effect of key hyperparameters in our framework. As shown in~\cref{fig:abl_k_alpha}, we analyze the interplay between the suppression strength $\alpha_0$ and $top\text{-}k$ in our CS-SVD module. A large $k$ leads to visible content leakage. Regarding $\alpha_0$, higher values apply stronger suppression but risk discarding fine-grained style textures. We also evaluate the robustness of SS-CFG to the guidance weight $w$. As illustrated in~\cref{fig:abl_sscfg_w}, our method maintains high visual quality even under large values of $w$, demonstrating strong resilience to prompt amplification. This allows for flexible tuning without causing content distortion or style degradation.

\section{Conclusion}
In this paper, we present~$\ours$, a diffusion-based method designed to mitigate the issue of content leakage in stylized image generation. We observe that when style embeddings are decomposed via SVD, the tail components often encode undesired content information. Building upon this insight, we propose CS-SVD, a time-aware soft exponential suppression strategy that effectively attenuates these tail components, significantly reducing content leakage.
To further enhance prompt fidelity and visual quality, we introduce the SS-CFG mechanism. Unlike the CFG formulation in existing encoder-based style transfer methods, SS-CFG constructs a style image-specific negative condition, enabling more precise suppression of undesired content guided by the characteristics of the style reference.
We integrate$~\ours$ with multiple representative style transfer frameworks and conduct comprehensive experiments. Both qualitative and quantitative results demonstrate that~$\ours$  effectively suppresses content leakage, enhances prompt fidelity, preserves rich stylistic details, and improves overall visual quality, making it a versatile and training-free solution for stylized image generation.

{
    \small
    \bibliographystyle{ieeenat_fullname}
    \bibliography{main}
}

\clearpage
\appendix
\newcommand{\AppendixPrefix}{A}
\clearpage
\setcounter{page}{1}
\setcounter{figure}{0}
\setcounter{table}{0}
\setcounter{equation}{0}

\renewcommand{\thefigure}{A.\arabic{figure}}
\renewcommand{\thetable}{A.\arabic{table}}
\renewcommand{\theequation}{A.\arabic{equation}}

\startcontents[appendix]
{
    \hypersetup{linkcolor=black}

    \titlecontents{lsection}[0pt]
        {\vspace{3pt}\bfseries}
        {\thecontentslabel\hspace{1em}}
        {}
        {\hfill\thecontentspage}
    \titlecontents{lsubsection}[1.5em]
        {\normalfont}
        {\thecontentslabel\hspace{1em}}
        {} 
        {\titlerule*[0.7pc]{.}\thecontentspage}
    \printcontents[appendix]{l}{1}{\section*{Appendix}}
}
\vspace{1em}

\section{More qualitative comparisons}
\begin{figure*}[t]
    \centering
    \includegraphics[width=1.0\linewidth]{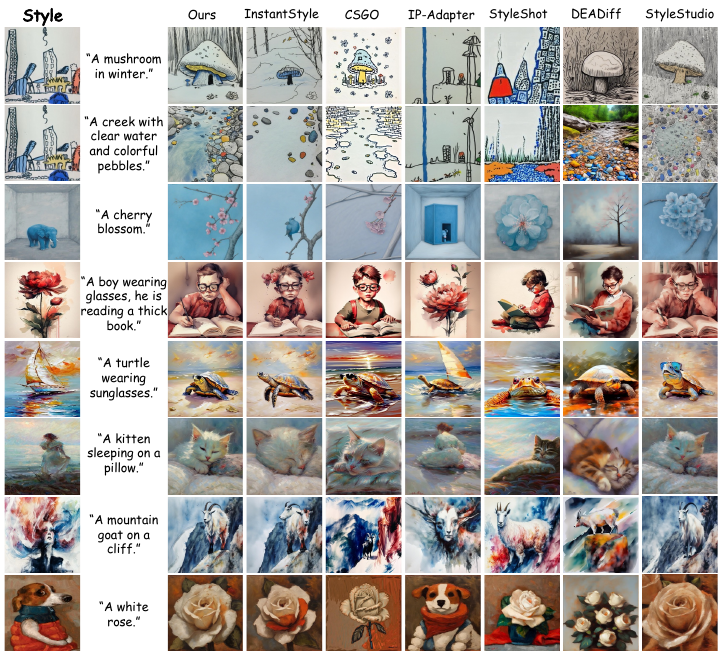}
    \caption{Qualitative comparison with the state-of-the-art encoder-based style transfer methods.}
    \label{fig:quality_4}
\end{figure*}

\begin{figure*}[t]
    \centering
    \includegraphics[width=1.0\linewidth]{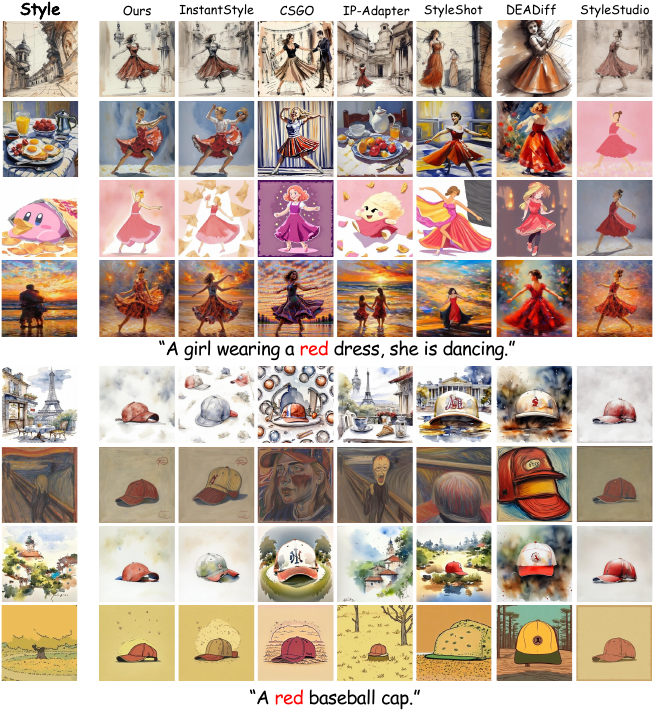}
    \caption{Qualitative comparison with the state-of-the-art encoder-based style transfer methods.}
    \label{fig:quality_1}
\end{figure*}

\begin{figure*}[t]
    \centering
    \includegraphics[width=1.0\linewidth]{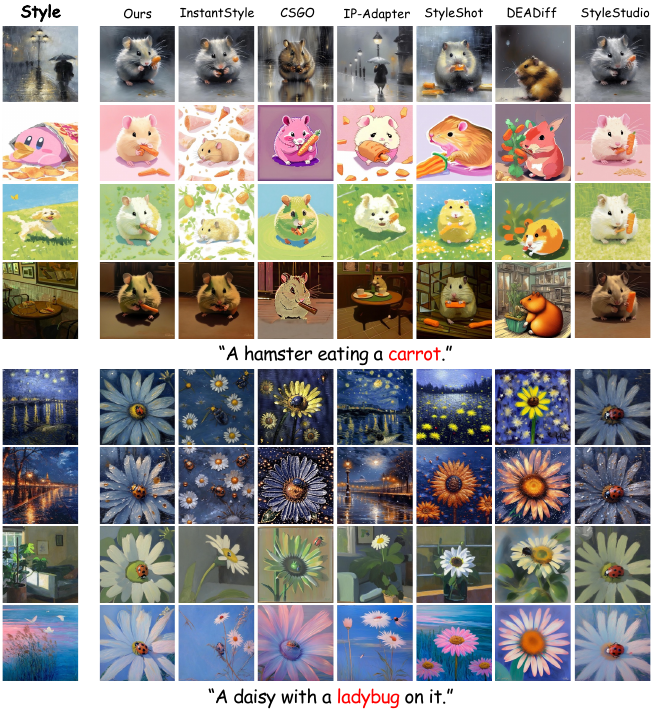}
    \caption{Qualitative comparison with the state-of-the-art encoder-based style transfer methods.}
    \label{fig:quality_2}
\end{figure*}

\begin{figure*}[t]
    \centering
    \includegraphics[width=1.0\linewidth]{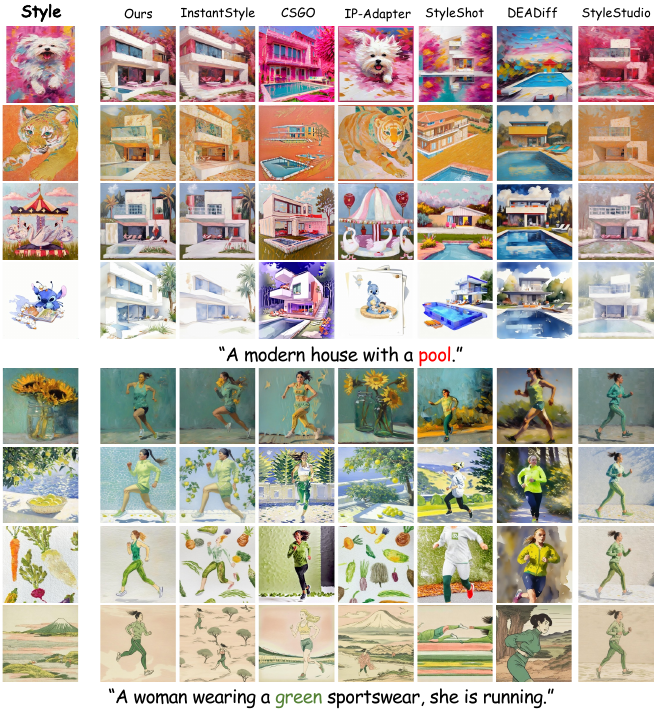}
    \caption{Qualitative comparison with the state-of-the-art encoder-based style transfer methods.}
    \label{fig:quality_3}
\end{figure*}

In this section, we provide more cases to compare our method with other state-of-the-arts (SOTAs). The models we choose are InstantStyle~\cite{wang2024instantstylefreelunchstylepreserving}, CSGO~\cite{xing2024csgocontentstylecompositiontexttoimage}, IP-Adapter~\cite{ye2023ipadaptertextcompatibleimage}, StyleShot~\cite{gao2025styleshotsnapshotstyle}, DEADiff~\cite{qi2024deadiffefficientstylizationdiffusion} and StyleStudio~\cite{lei2025stylestudio}.~\cref{fig:quality_1,fig:quality_2,fig:quality_3,fig:quality_4} are the results.

The qualitative comparison demonstrates that our approach consistently outperforms existing methods across various models, achieving better text alignment and reduced content leakage. Moreover, it effectively preserves stylistic attributes such as color and texture. These results highlight both the effectiveness and robustness of our method. In other words, the method not only produces outputs that are more faithful to the intended semantics, but also maintains high visual fidelity in terms of stylistic consistency. Such consistent improvements across different conditions further validate the effectiveness and generalizability of our approach.

\section{Quantitative comparison on StyleBench~\cite{gao2025styleshotsnapshotstyle}}
As a complement to the $\ours$ benchmark results in the main paper, we report additional quantitative comparisons on the StyleBench dataset~\cite{gao2025styleshotsnapshotstyle}, which includes 490 diverse style images and 20 standardized prompts. The results are presented in~\cref{tab:quantitative_stylebench}. Across different base models, our method consistently improves CLIP Text Alignment (CLIP TA)~\cite{radford2021clip}, indicating better prompt compliance. While slight fluctuations are observed in style similarity metrics (CLIP SS~\cite{radford2021clip}, DINO SS~\cite{caron2021dino}), this aligns with our goal of reducing style-induced content leakage, which may affect semantic-heavy evaluations. These results further demonstrate the general applicability and robustness of our method across architectures.

\begin{figure*}[t]
    \centering
    \includegraphics[width=1.0\linewidth]{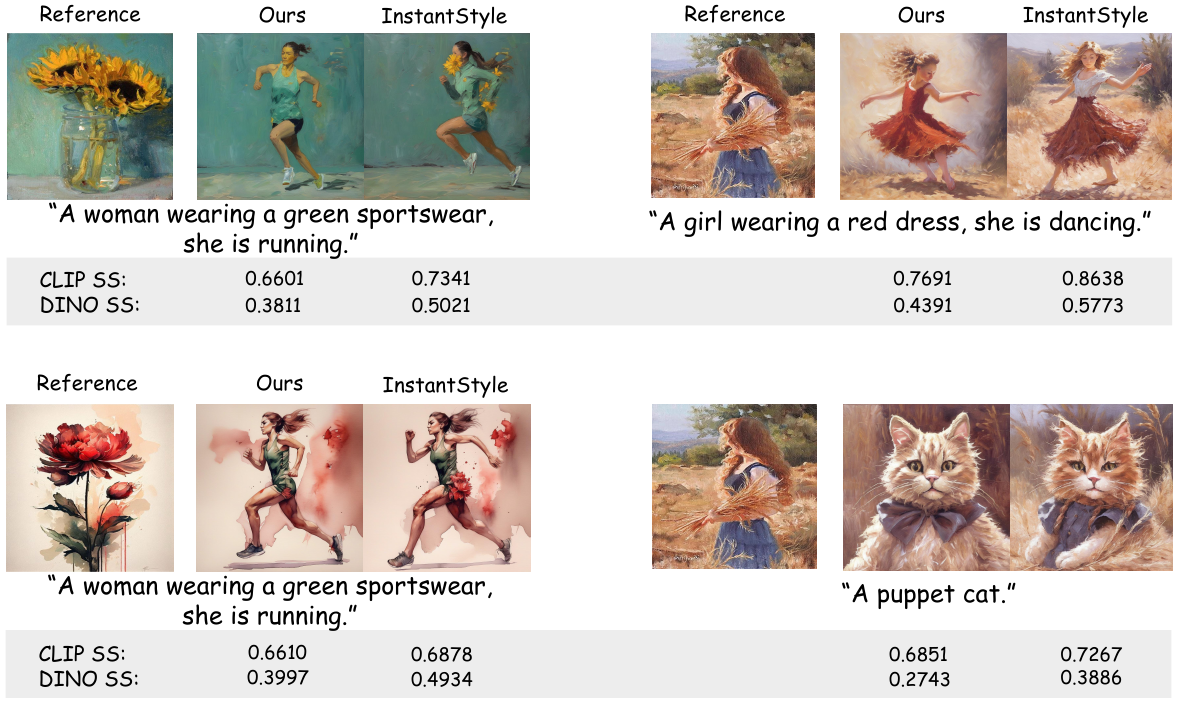}
    \caption{Style similarity comparison between our method and baselines with visible content leakage. We observe that instances with content leakage (\eg, extraneous flowers, backgrounds, or accessories copied from the style reference) often yield higher CLIP-SS~\cite{radford2021clip} and DINO-SS~\cite{caron2021dino} scores, despite compromising prompt fidelity and visual coherence. This highlights a potential limitation of current style similarity metrics, which may overestimate stylization quality when semantic content from the style image is inadvertently transferred.}
    \label{fig:ss_compare}
\end{figure*}

While CLIP-SS~\cite{radford2021clip} and DINO-SS~\cite{caron2021dino} are widely used as style similarity metrics, they do not explicitly penalize content leakage. In some case~\cref{fig:ss_compare}, we observe that results containing semantic elements from the reference image can still achieve high style similarity scores, despite violating the intended prompt semantics.
To further examine this phenomenon, we compare results that exhibit content leakage with those generated by our method. As shown in~\cref{fig:ss_compare}, although our approach yields slightly lower style similarity scores, it clearly suppresses undesired content (\eg, wheat fields, puppets) and better aligns with the target prompts. This suggests that existing style metrics may conflate content with style and overlook violations of prompt fidelity.
This analysis highlights a potential limitation of current metrics and reinforces the importance of human-perceivable alignment in stylized generation.

\section{Impact of parameters}
To further investigate the effect of key hyperparameters in CS-SVD, we conduct a grid search over the suppression strength $\alpha_0$ and the truncation rank $k$. 
Due to computational constraints, the evaluation is performed on a representative subset of our $\ours$ benchmark, consisting of 5 randomly selected prompts and the full set of 100 style images.
We report three commonly used quantitative metrics, CLIP Text Alignment (CLIP-TA)~\cite{radford2021clip}, CLIP Style Similarity (CLIP-SS)~\cite{radford2021clip}, and DINO Style Similarity (DINO-SS)~\cite{caron2021dino}, under each $(k, \alpha_0)$ configuration.

The corresponding quantitative results are presented in~\cref{tab:hyper_all}. We observe that smaller values of $k$ generally yield higher CLIP-TA~\cite{radford2021clip} scores, likely due to stronger suppression of content leakage. However, this often comes at the cost of losing meaningful stylistic textures, as further illustrated in the subsequent qualitative results. 
On the other hand, larger values of $k$ tend to retain more embedding components, some of which may be associated with semantic noise rather than beneficial style features, leading to higher style similarity scores but reduced prompt alignment and increased content leakage. 
The suppression factor $\alpha_0$ also plays a key role in adjusting the filtering strength. While increasing $\alpha_0$ can attenuate more residual content signals, overly large values may also wash out essential stylistic cues. These observations underscore the necessity of striking a balanced configuration that mitigates content leakage while maintaining stylistic fidelity, without requiring exhaustive tuning.

\begin{table}[t]
\centering
\begin{tabular}{lcccc}
\toprule
Method&CLIP TA$\uparrow$&CLIP SS$\uparrow$ &DINO SS$\uparrow$\\
\midrule
\cellcolor{lightCyan}\textbf{InstantStyle$^\dag$} & \cellcolor{lightCyan}\textbf{0.231} & \cellcolor{lightCyan}\textbf{0.640} & \cellcolor{lightCyan}\textbf{0.335}\\
InstantStyle & 0.227 & 0.665 & 0.386\\
\midrule
\cellcolor{lightCyan}\textbf{DEADiff$^\dag$} & \cellcolor{lightCyan}\textbf{0.234} & \cellcolor{lightCyan}\textbf{0.602} & \cellcolor{lightCyan}\textbf{0.306}\\  
DEADiff & 0.231 & 0.619 & 0.330\\
\midrule
\cellcolor{lightCyan}\textbf{StyleShot$^\dag$} & \cellcolor{lightCyan}\textbf{0.222} & \cellcolor{lightCyan}\textbf{0.661} & \cellcolor{lightCyan}\textbf{0.411}\\
StyleShot & 0.215 & 0.677 & 0.456\\
\midrule
StyleStudio & 0.230 & 0.672 & 0.402\\
CSGO & 0.223 & 0.608 & 0.386\\
IP-Adapter & 0.137 & 0.829 & 0.661\\
\bottomrule
\end{tabular}
\caption{Quantitative Comparisons on StyleBench. $\dag$: Baseline integrated with our method.}
\label{tab:quantitative_stylebench}
\end{table}

\section{Integration with other methods}
\subsection{Qualitative results on StyleShot~\cite{gao2025styleshotsnapshotstyle}}
\begin{figure*}[t]
    \centering
    \includegraphics[width=1.0\linewidth]{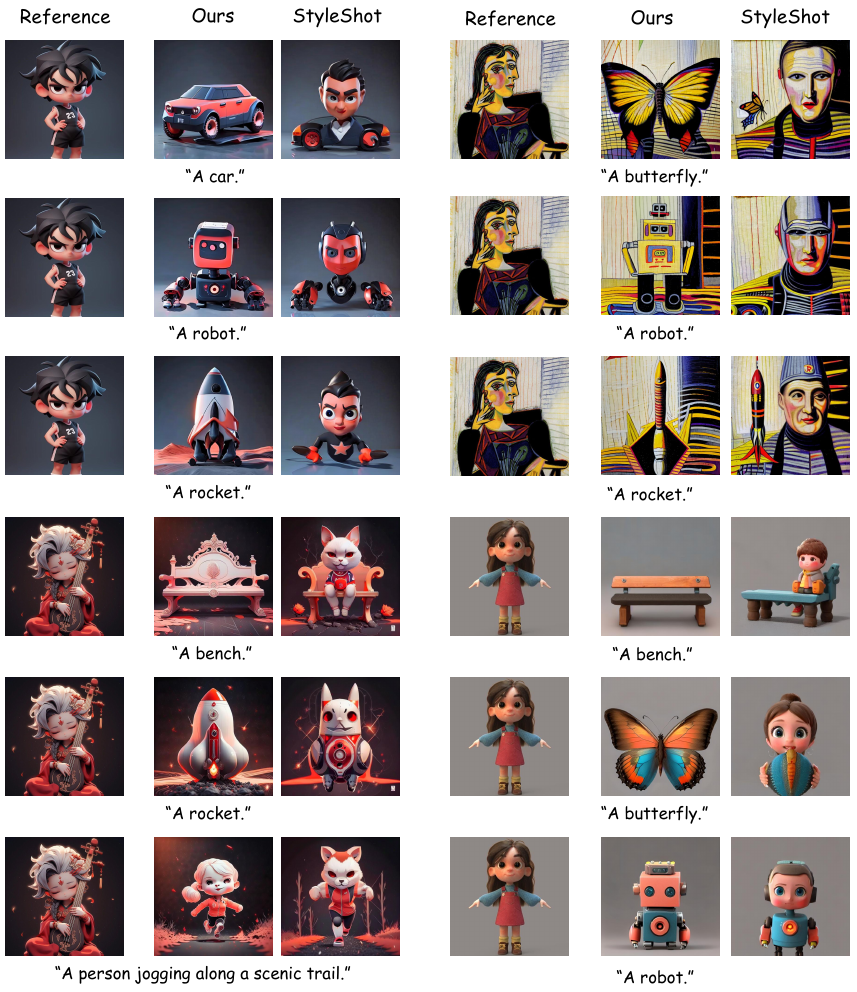}
    \caption{The qualitative comparison between ours (Combined with StyleShot) and StyleShot~\cite{gao2025styleshotsnapshotstyle}.}
    \label{fig:styleshot3}
\end{figure*}

\begin{figure*}[t]
    \centering
    \includegraphics[width=1.0\linewidth]{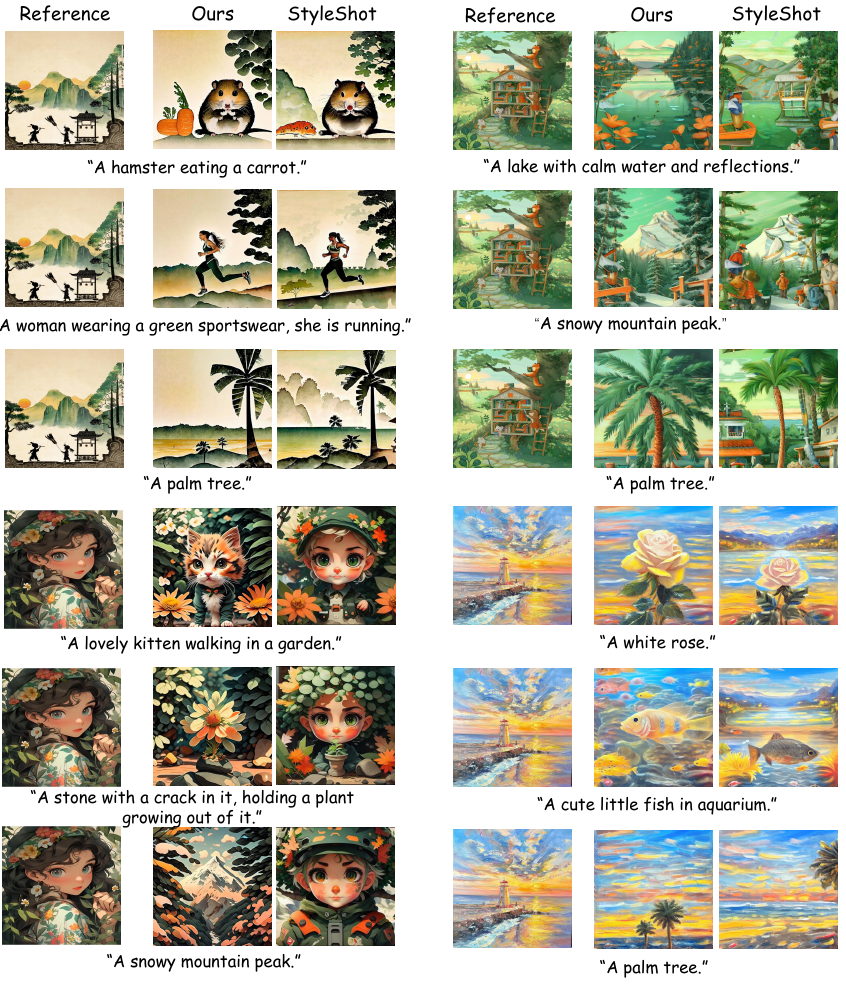}
    \caption{The qualitative comparison between ours (Combined with StyleShot) and StyleShot~\cite{gao2025styleshotsnapshotstyle}. In the case of ``A hamster eating a carrot", the mountain in the background is the leakage content. The leakage content of ``A lake with calm water and reflecting." is the wood shelf.}
    \label{fig:styleshot1}
\end{figure*}

\begin{figure*}[t]
    \centering
    \includegraphics[width=1.0\linewidth]{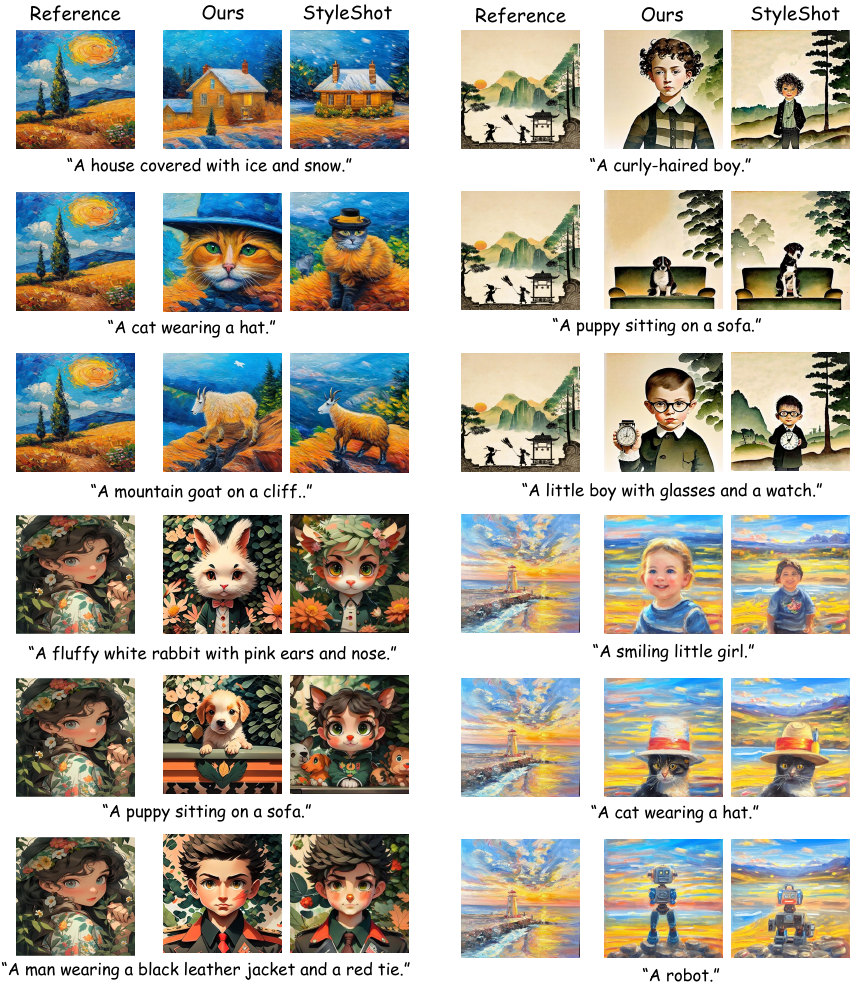}
    \caption{The qualitative comparison between ours (Combined with StyleShot) and StyleShot~\cite{gao2025styleshotsnapshotstyle}. The leakage content of ``A house covered with ice and snow." is green tree. The leakage content of ``A mountain goat on a cliff." is the yellow grass.}
    \label{fig:styleshot2}
\end{figure*}

In this section, we integrate our approach with StyleShot~\cite{gao2025styleshotsnapshotstyle}. It is based on SD1.5, and also use encoder to extract style features. The settings and performing process of CS-SVD and SS-CFG are the same as on InstantStyle. ~\cref{fig:styleshot3}, ~\cref{fig:styleshot1}, and ~\cref{fig:styleshot2} are the qualitative results.

On the StyleShot~\cite{gao2025styleshotsnapshotstyle}, our method demonstrates clear advantages over the baseline. Specifically, it not only achieves better text alignment and stylistic consistency, but also excels at preserving fine-grained visual attributes such as color and texture. 
One of the key improvements lies in mitigating content leakage, where semantically irrelevant or misleading visual elements appear in the generated images.

As shown in ~\cref{fig:styleshot1}, for the prompts ``A lovely kitten walking in a garden", ``A stone with a crack in it, holding a plant growing out of it", and ``A snowy mountain peak", the baseline methods undesirably introduce human faces, the content that is semantically unrelated and stylistically inconsistent. Our method effectively suppresses this leakage while preserving the intended style. 
Similarly, in the prompts ``A lake with calm water and reflections" and ``A palm tree", the baseline models generate unwanted visual elements such as wooden shelves, which disrupt the scene understanding. In contrast, our approach avoids these artifacts while maintaining stylistic fidelity. In~\cref{fig:styleshot2}, the prompt ``A house covered with ice and snow" suffers from content leakage in the form of a green tree, contradicting the winter theme conveyed by the text. Our method corrects this inconsistency by aligning visual output with commonsense semantics. 
Finally, as shown in~\cref{fig:styleshot3}, spurious character faces that appear in the baseline generations are successfully filtered out by our model, demonstrating its effectiveness in eliminating undesired content while preserving stylistic expressiveness.

This consistent improvement on StyleShot~\cite{gao2025styleshotsnapshotstyle} further supports the generality of our approach, indicating that the method is not limited to a specific setting but can be reliably applied across different styles and conditions. Such results once again highlight the robustness and effectiveness of our framework in handling challenging style transfer scenarios.

\subsection{Qualitative results on DEADiff~\cite{qi2024deadiffefficientstylizationdiffusion}}
\begin{figure*}[t]
    \centering
    \includegraphics[width=1.0\linewidth]{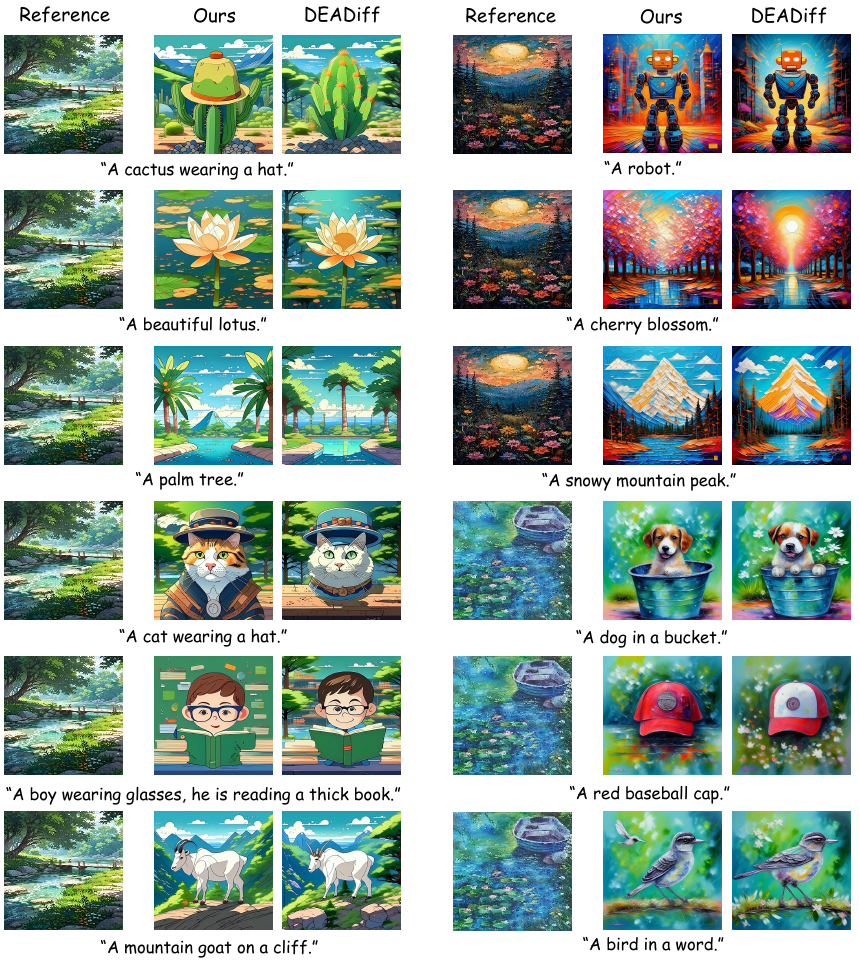}
    \caption{The qualitative comparison between ours (Combined with DEADiff) and DEADiff~\cite{qi2024deadiffefficientstylizationdiffusion}. In the case of ``A beautiful latus.", the lotus leaves in the pond have turned into leaves. In the case of ``A palm tree.", the leaves of the palm tree have become ordinary leaves and lost their characteristics.}
    \label{fig:deadiff_appendix1}
\end{figure*}

\begin{figure*}[t]
    \centering
    \includegraphics[width=1.0\linewidth]{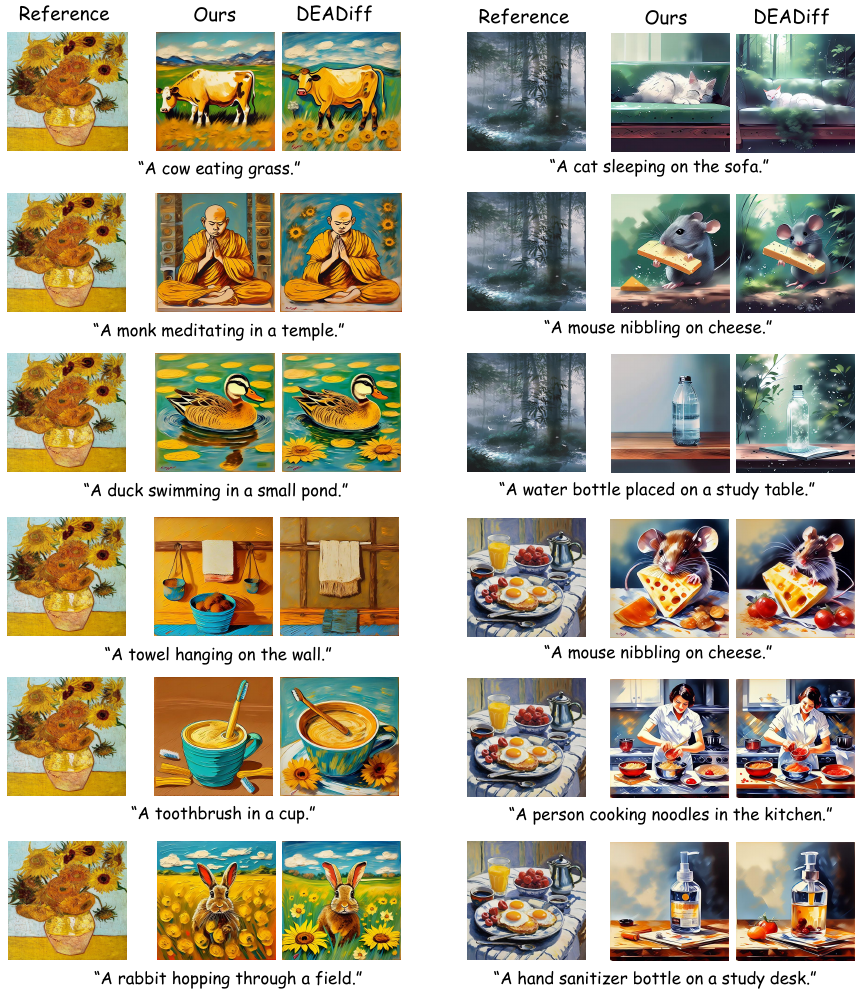}
    \caption{The qualitative comparison between ours (Combined with DEADiff) and DEADiff~\cite{qi2024deadiffefficientstylizationdiffusion}. In the case of ``A duck swimming in a small pond.", the floating duckweed in the pond turned into yellow flowers. In the case of ``A person cooking noodles in the kitchen.", the noodles in the pot and bowl turned into red berries.}
    \label{fig:deadiff_appendix2}
\end{figure*}

To verify the generalization of our method, we combine it with DEADiff~\cite{qi2024deadiffefficientstylizationdiffusion}.
The qualitative results are shown in~\cref{fig:deadiff_appendix1}, ~\cref{fig:deadiff_appendix2}. As shown in~\cref{fig:deadiff_appendix1}, for the prompt ``A beautiful lotus.”, the result generated by DEADiff~\cite{qi2024deadiffefficientstylizationdiffusion} includes unrelated green leaves under the lotus, rather than actual lotus leaves, which disrupts semantic alignment with the input text. In contrast, our method successfully mitigates this content leakage while preserving stylistic integrity. Similarly, for the prompt ``A palm tree.”, DEADiff~\cite{qi2024deadiffefficientstylizationdiffusion} renders generic leaves that fail to capture the distinct features of palm fronds, whereas our approach maintains these stylistic details faithfully.
In~\cref{fig:deadiff_appendix2}, for the case ``A duck swimming in a small pond.”, DEADiff~\cite{qi2024deadiffefficientstylizationdiffusion} introduces yellow flowers in place of duckweed, deviating from the original content. For the prompt ``A person cooking noodles in the kitchen.”, the noodles are erroneously depicted as red berries by DEADiff~\cite{qi2024deadiffefficientstylizationdiffusion}. Our method effectively corrects such content inaccuracies, ensuring text-image consistency while retaining the artistic style. In the case of ``A cat sleeping on the sofa.”, ``A mouse nibbling on cheese.” and ``A water bottle placed on a study table.”, the unwanted bamboo leaves are suppressed by our approach.
These results demonstrate our method’s advantage in addressing content leakage issues inherent to DEADiff~\cite{qi2024deadiffefficientstylizationdiffusion} without destroying stylistic details. 

\section{Details of the user study}
\begin{figure*}[t]
    \centering
    \includegraphics[width=1.0\linewidth]{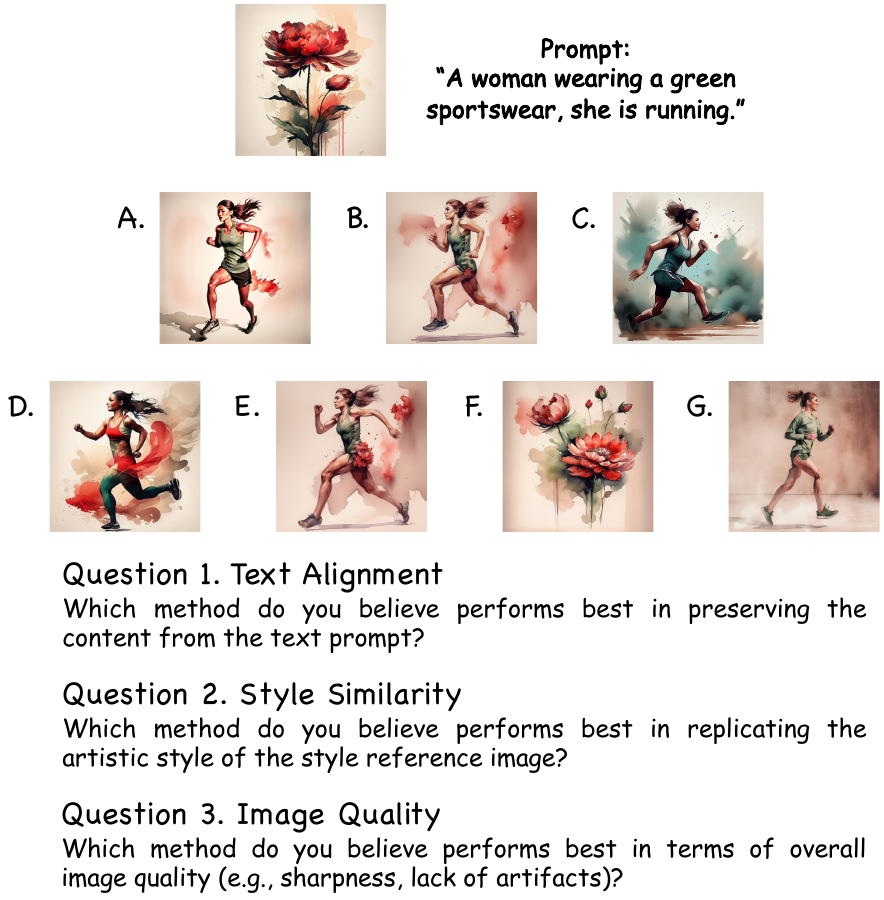}
    \caption{The questionnaire format for the user study. Each option represents the generation result of a method under a given style and prompt.}
    \label{fig:userstudy}
\end{figure*}

In this section, we present the questionnaire format used in our user study. 
Because stylized generation is inherently subjective, human evaluation is essential for a reliable assessment of perceptual quality. 
Our study consists of 60 questions derived from 20 style–prompt pairs, each evaluated along three criteria: text alignment, style similarity, and overall image quality. 
For every question, participants were shown outputs from all methods in a randomized order and asked to choose the best result under the specified criterion. 
We collected 43 complete responses, yielding a total of 2,580 individual judgments, which provides broad and diverse human preference coverage. 
The questionnaire format is illustrated in~\cref{fig:userstudy}.

\begin{table*}[t]
\centering
\renewcommand{\arraystretch}{1.10}
\setlength{\tabcolsep}{6pt}  %
\scriptsize
\begin{tabular}{
c
cccc
cccc
cccc
}
\toprule
& \multicolumn{4}{c}{\textbf{CLIP-TA$\uparrow$~\cite{radford2021clip}}} 
& \multicolumn{4}{c}{\textbf{CLIP-SS$\uparrow$~\cite{radford2021clip}}}
& \multicolumn{4}{c}{\textbf{DINO-SS$\uparrow$~\cite{caron2021dino}}} \\
\cmidrule(r){2-5} \cmidrule(r){6-9} \cmidrule(){10-13}
\textbf{}
& $\alpha_0$=0.01 & $\alpha_0$=0.02 & $\alpha_0$=0.03 & $\alpha_0$=0.05
& $\alpha_0$=0.01 & $\alpha_0$=0.02 & $\alpha_0$=0.03 & $\alpha_0$=0.05
& $\alpha_0$=0.01 & $\alpha_0$=0.02 & $\alpha_0$=0.03 & $\alpha_0$=0.05\\
\midrule
k=0 & 0.241 & 0.250 & 0.251 & 0.251
  & 0.598 & 0.576 & 0.566 & 0.557
  & 0.317 & 0.265 & 0.240 & 0.222 \\
k=1 & 0.230 & 0.235 & 0.237 & 0.239
  & 0.625 & 0.612 & 0.606 & 0.599
  & 0.382 & 0.359 & 0.348 & 0.333 \\
k=2 & 0.223 & 0.224 & 0.225 & 0.226
  & 0.641 & 0.639 & 0.637 & 0.635
  & 0.412 & 0.408 & 0.406 & 0.403 \\
k=3 & 0.222 & 0.222 & 0.223 & 0.224
  & 0.642 & 0.641 & 0.641 & 0.640
  & 0.412 & 0.412 & 0.411 & 0.410 \\
\bottomrule
\end{tabular}

\vspace{2mm}
\caption{
Hyperparameter study of SVD truncation rank $k$ and suppression strength $\alpha_0$ across
CLIP-TA~\cite{radford2021clip}, CLIP-SS~\cite{radford2021clip}, and DINO-SS~\cite{caron2021dino}. 
Larger $k$ preserves more singular components but reintroduces content leakage, while larger $\alpha_0$ applies stronger suppression and may weaken style richness.
A balanced trade-off is achieved around $k{=}1$ and $\alpha_0{\in}[0.01,\,0.02]$.
}
\label{tab:hyper_all}
\end{table*}

\section{Details of the $\ours$ dataset}
\begin{figure*}[t]
    \centering
    \includegraphics[width=1.0\linewidth]{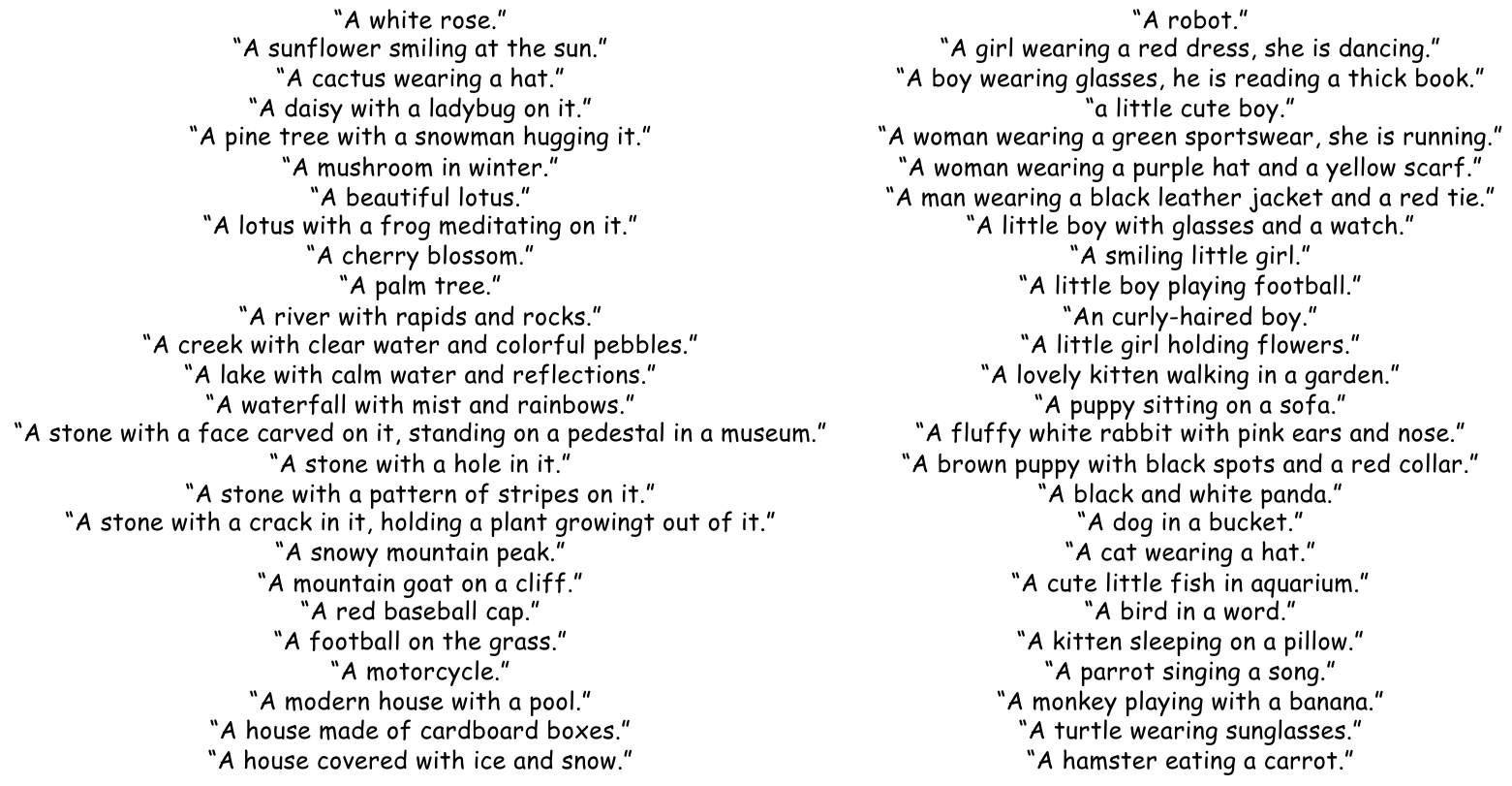}
    \caption{The prompts used in the quantitative experiments were derived from StyleAdapter~\cite{wang2024styleadapterunifiedstylizedimage}.}
    \label{fig:dataset_styleadapter}
\end{figure*}

The $\ours$ dataset is constructed by exhaustively pairing 52 text prompts from StyleAdapter~\cite{wang2024styleadapterunifiedstylizedimage} with 100 curated style images, resulting in 5,200 text-style pairs. The 52 prompts originate from a widely adopted benchmark in stylization research and are specifically designed to cover a broad range of semantic structures, including objects, scenes, and multi-clause descriptions. These prompts are illustrated in~\cref{fig:dataset_styleadapter}.

\begin{figure*}[t]
    \centering
    \includegraphics[width=1.0\linewidth]{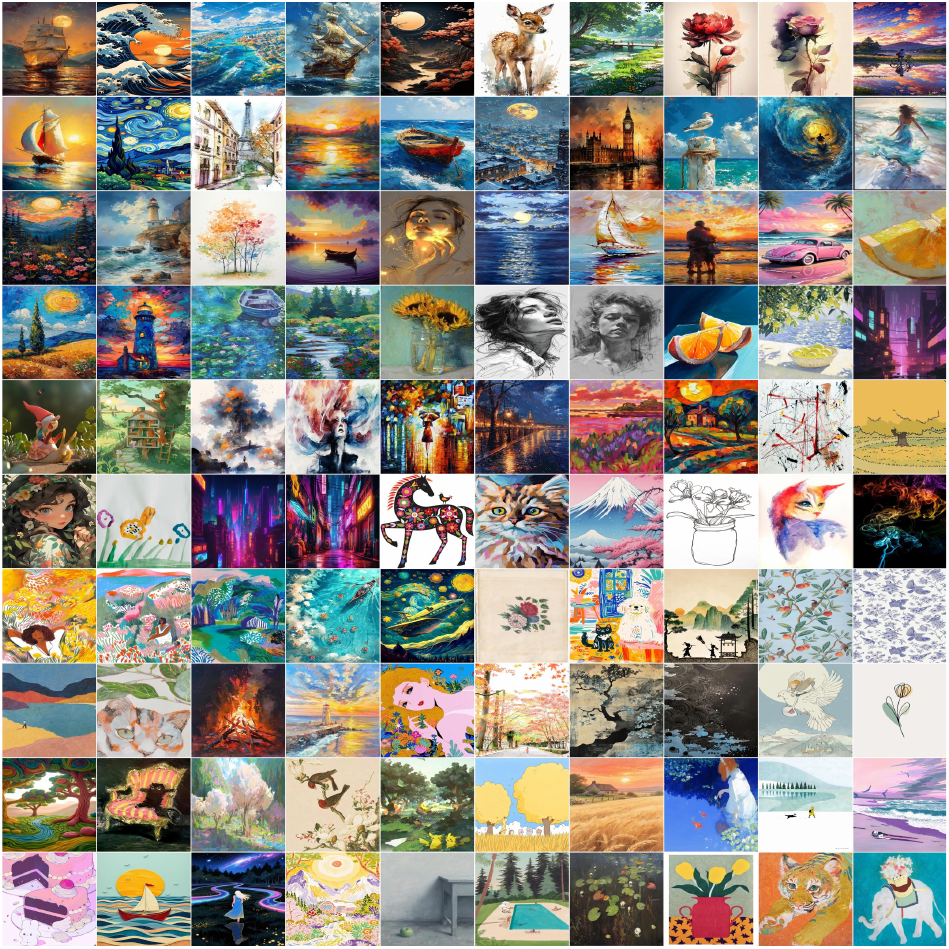}
    \caption{The styles used for quantitative comparisons in $\ours$.}
    \label{fig:dataset_CleanStyle}
\end{figure*}

To complement this textual diversity, we collected 100 representative style images spanning a wide range of artistic genres, color palettes, and texture characteristics, as shown in~\cref{fig:dataset_CleanStyle}. By pairing every prompt with every style reference, the $\ours$ dataset offers a systematic and comprehensive evaluation suite, enabling controlled assessment of both text adherence and stylistic consistency across diverse generation conditions.

\section{Limitation and future work}
The main limitation of $\ours$ is the additional inference cost introduced by applying CS-SVD throughout the denoising process. Although the size of the style-related Key/Value matrices influences runtime, the dominant factor is the number of cross-attention layers to which CS-SVD is applied and the total number of diffusion iterations. Because SVD is executed repeatedly across timesteps, the overhead accumulates proportionally with both layer depth and sampling length. Future improvements may explore approximate or cached decompositions to further reduce this computational burden.

A natural extension of this work is to adapt our analytical filtering and style-aware guidance framework to image-to-image style transfer. Since the core mechanism operates directly on style embeddings rather than text conditioning, $\ours$ is well aligned with I2I pipelines, and extending it to these settings may offer more controllable and content-preserving stylization across broader tasks.



\end{document}